\documentclass[11pt,letterpaper]{article}

\usepackage[letterpaper,margin=1in]{geometry}
\usepackage{amsmath}
\usepackage{amssymb}
\usepackage{booktabs}
\usepackage{array}
\usepackage{tabularx}
\usepackage{graphicx}
\usepackage{xcolor}
\usepackage{url}
\usepackage[numbers,sort&compress]{natbib}

\usepackage{placeins}
\usepackage{float}
\usepackage{hyperref}
\usepackage{tikz}
\usetikzlibrary{arrows.meta,positioning,fit,calc}


\hypersetup{
    colorlinks=true,
    linkcolor=blue,
    citecolor=blue,
    urlcolor=blue,
    pdftitle={Generative Atmospheric Super-Resolution from Heterogeneous In Situ Observations through Composable Interfaces},
    pdfauthor={Yang Xu, Dibyajyoti Chakraborty, Haiwen Guan, Sen Wang, and Romit Maulik}
}

\graphicspath{{figures/}}

\newcommand{\R}{\mathbb{R}}

\newcommand{\Lobs}{\mathcal{L}_{\mathrm{obs}}}

\newcommand{\Higra}{H_{\mathrm{R}}}

\newcommand{\Mset}{\mathcal{M}}
\newcommand{\Nens}{N_{\mathrm{ens}}}
\newcommand{\xclean}{\mathbf{x}_0}
\newcommand{\xnoisy}{\mathbf{x}_{\sigma}}
\newcommand{\xden}{\widehat{\mathbf{x}}_0}

\title{
Generative Atmospheric Super-Resolution from Heterogeneous In~Situ Observations through Composable Interfaces
}

\author{
Yang Xu$^{1}$ \and
Dibyajyoti Chakraborty$^{2}$ \and
Haiwen Guan$^{2}$ \and
Sen Wang$^{1}$ \and
Romit Maulik$^{1}$
}

\date{
$^{1}$School of Mechanical Engineering, Purdue University, West Lafayette, Indiana, USA \\
$^{2}$College of Information Sciences and Technology, The Pennsylvania State University, University Park, Pennsylvania, USA}

\begin{document}
\maketitle

\begin{abstract}
Atmospheric observations are sparse, heterogeneous, and unevenly distributed, 
whereas many generative atmospheric models learn 
distributions over regularly gridded multivariate states. Once pretrained, diffusion
models can supply atmospheric priors that can be combined with observation-derived likelihood factors in a
Bayesian formulation.
However, these observation sources differ substantially in
geometry and sampling density, complicating the consistent use of their
observations within a common inference framework.
Here, we formulate this reconstruction problem as generative atmospheric super-resolution and introduce composable observation interfaces for conditioning a single pretrained 13-variable atmospheric diffusion model.
The interfaces convert sparse radiosonde (R), clustered
aircraft (A), and dense irregular surface-station (S) observations into
source-specific likelihood factors that specify where observations constrain
the gridded state, how residuals are counted under uneven sampling, and how
strongly each source guides posterior sampling.
We developed the aircraft and surface observation interfaces
using 2019 observations and evaluated the selected interfaces throughout 2020
without further tuning.
Compared with reconstructions conditioned only on radiosonde
observations, the composed R+A+S interface reduces RMSE evaluated against ERA5
by $9.24\%$ across all 13 state variables over the CONUS domain.
The aircraft and surface factors provide complementary
improvements in upper-air and surface variables.
The R+A+S combination also lowers the Continuous Ranked
Probability Score (CRPS), while evaluations at held-out aircraft and
surface-station observations show reduced prediction errors.
Together, these results demonstrate a modular route for
conditioning a pretrained atmospheric generative prior on heterogeneous in
situ observations without retraining the underlying model.

\end{abstract}

\section{Introduction}

The atmosphere is a high-dimensional, multiscale system whose state is partially observed \citep{carrassi2018dataassimilation}.
These observations are collected by instruments and networks that differ in measured variables, spatial and vertical support, and sampling patterns.
Radiosonde networks provide spatially sparse vertical profiles, aircraft measurements are collected in flight, and surface-station networks provide measurements at fixed locations \citep{durre2018igra,wagner2021airborne,noaa_madis_surface}.
By contrast, several atmospheric generative models learn distributions over regularly gridded reanalysis states \citep{dj_diffusion_prior,huang2024diffda,price2025gencast}.
We frame the reconstruction of a regularly gridded atmospheric
state from sparse, heterogeneous, and irregularly sampled observations as generative
atmospheric super-resolution.
In a Bayesian formulation, a pretrained atmospheric diffusion
model supplies a prior over these gridded states, while measurements from each
observation source enter a likelihood that relates those measurements to a
candidate state
\citep{carrassi2018dataassimilation,chung2023dps}.
The pretrained prior models spatial and cross-variable statistical relationships within the atmospheric state \citep{dj_diffusion_prior}, but it does not by itself specify how observations with different geometries and sampling densities should enter the likelihood.
Consistently defining and combining likelihood factors across
these in situ observation sources is therefore the Bayesian problem considered
here.

We define an \emph{observation interface} as the set of source-specific choices that connects measurements to the gridded atmospheric state.
Each interface specifies which measurements are retained, how they are mapped to state variables and locations, how their residuals are counted, and how the resulting likelihood factor is weighted during inference.
For aircraft and surface-station observations, reports within each model-grid cell are averaged separately for each variable before evaluating the likelihood, so densely reported cells do not receive greater weight solely because they contain more reports.

Techniques for solving inverse problems with diffusion models can incorporate new measurements during reconstruction without retraining the pretrained model for each inverse problem \citep{song2022medical,graikos2022plugandplay,kawar2022ddrm,chung2023dps,song2023pigdm}.
Diffusion posterior sampling (DPS) combines the pretrained prior with a measurement likelihood during sampling to approximate the corresponding posterior distribution \citep{chung2023dps}.
In our setting, this allows each source-specific observation interface to connect its observations to the same pretrained atmospheric diffusion model.

Generative approaches to atmospheric state reconstruction now span inference-time guidance, learned multimodal conditioning, direct observation-to-analysis mappings, and models that operate in observation space \citep{rozet2023score,huang2024diffda,andry2025appa,qu2024slams,allen2025aardvark,gholoubi2026ocelot}.
The direct predecessor to this study developed the 13-variable ERA5 diffusion prior and stochastic guidance sampler used here and demonstrated zero-shot conditioning on coarse ERA5 fields, IGRA radiosondes, and atmospheric-emulator outputs \citep{dj_diffusion_prior}. We retain this pretrained prior and sampler while shifting the methodological focus to constructing likelihood factors from heterogeneous observation sources. 
The question addressed here is how observations from
structurally different in situ sources can be represented as explicit and
composable likelihood factors for a single pretrained atmospheric prior over
gridded states without retraining the prior.

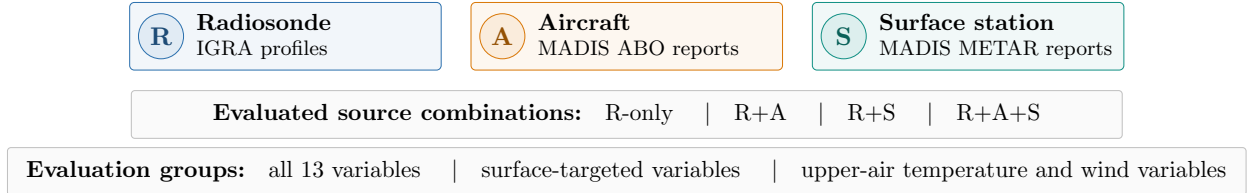
\begin{figure}[H]
    \centering
    \resizebox{\textwidth}{!}{\begin{tikzpicture}[
    font=\small,
    family/.style={draw=#1, fill=#1!6, rounded corners=2pt,
                   minimum width=4.70cm, minimum height=1.02cm,
                   inner sep=0pt},
    badge/.style={circle, draw=#1, fill=#1!14, text=#1!68!black,
                  minimum size=0.68cm, inner sep=0pt, font=\bfseries},
    note/.style={draw=black!25, fill=black!2, rounded corners=1.5pt,
                 minimum width=14.95cm, minimum height=0.72cm,
                 align=center, inner xsep=8pt}
]
\definecolor{radiosondeblue}{HTML}{2A6FAD}
\definecolor{aircraftorange}{HTML}{D97706}
\definecolor{surfacegreen}{HTML}{0F8F83}

\node[family=radiosondeblue] (rbox) at (0,0) {};
\node[badge=radiosondeblue] at ([xshift=0.48cm]rbox.west) {R};
\node[anchor=west, align=left] at ([xshift=0.88cm]rbox.west)
  {\textbf{Radiosonde}\\[-1pt]{\footnotesize IGRA profiles}};

\node[family=aircraftorange] (abox) at (5.15,0) {};
\node[badge=aircraftorange] at ([xshift=0.48cm]abox.west) {A};
\node[anchor=west, align=left] at ([xshift=0.88cm]abox.west)
  {\textbf{Aircraft}\\[-1pt]{\footnotesize MADIS ABO reports}};

\node[family=surfacegreen] (sbox) at (10.30,0) {};
\node[badge=surfacegreen] at ([xshift=0.48cm]sbox.west) {S};
\node[anchor=west, align=left] at ([xshift=0.88cm]sbox.west)
  {\textbf{Surface station}\\[-1pt]{\footnotesize MADIS METAR reports}};

\node[note] at (5.15,-1.18)
  {\textbf{Evaluated source combinations:}\quad
   R-only \quad\textbar\quad R+A \quad\textbar\quad
   R+S \quad\textbar\quad R+A+S};

\node[note] at (5.15,-2.04)
  {\textbf{Evaluation groups:}\quad
   all 13 variables \quad\textbar\quad surface-targeted variables
   \quad\textbar\quad upper-air temperature and wind variables};
\end{tikzpicture}}
    \caption{Observation sources, evaluated source combinations, and evaluation groups used throughout the study. The surface-targeted group and
    upper-air temperature and wind group contain the state variables directly constrained by S and A, respectively.}
    \label{fig:terminology_map}
\end{figure}

Each reconstruction targets a single analysis time.
We study three observation sources: radiosonde observations
(R) from the Integrated Global Radiosonde Archive (IGRA); aircraft
observations (A) from the Aircraft Based Observations (ABO) product of the
National Oceanic and Atmospheric Administration (NOAA) Meteorological
Assimilation Data Ingest System (MADIS); and surface-station observations (S)
from the MADIS Meteorological Aerodrome Report (METAR) product
\citep{noaa_ncei_igra,noaa_madis_aircraft,noaa_madis_surface}.
Figure~\ref{fig:terminology_map} introduces the R/A/S notation, the evaluated
source combinations, and the variable groups used below.
R can constrain all 13 state variables globally where corresponding measurements are available.
We define the contiguous United States (CONUS) domain as the fixed rectangular computational domain spanning $24^{\circ}$--$50^{\circ}$N and $125^{\circ}$--$66^{\circ}$W, which includes adjacent portions of southern Canada, northern Mexico, and surrounding waters.
Within the CONUS domain, A constrains temperature and horizontal winds at 500 and 850 hPa, while S constrains 2 m temperature and 10 m winds.
Each source retains its own observation construction, state mapping, and residual-counting rule as the likelihood factors are composed in the R-only, R+A, R+S, and R+A+S configurations with the same pretrained prior.

We selected the A and S interface designs and calibrated their
likelihood parameters using 24 prespecified development cases from 2019, then
evaluated the selected interfaces at 723 matched analysis times in 2020
without further tuning. Over the CONUS domain, R+A+S reduces ERA5-referenced RMSE
relative to R-only conditioning for all 13 state variables, while
probabilistic metrics and held-out evaluations provide additional evidence of
complementary surface and upper-air gains.

Methods for relating observations to model states and for averaging densely sampled observations are established in classical data assimilation \citep{carrassi2018dataassimilation,janjic2018representation}, while DPS incorporates a measurement likelihood during sampling \citep{chung2023dps}.

Our contribution is to organize and evaluate these choices as
explicit, auditable, source-specific observation interfaces that allow one
pretrained atmospheric prior to be reused across heterogeneous in situ
observation sources.

This contribution has two parts:

\begin{enumerate}

\item \textbf{Source-specific observation interfaces.} We
define how measurements from each source are selected, mapped to the model
grid, aggregated, and weighted in the likelihood
(Section~\ref{sec:observation_factorization}). Their separate effects are
evaluated through annual gridded comparisons and held-out observation
evaluations (Sections~\ref{sec:annual_rmse_results}
and~\ref{sec:heldout_results}).

\item \textbf{Inference-time composition with a fixed prior.}
We combine the R, A, and S likelihood factors around one pretrained
13-variable atmospheric diffusion model without retraining it when the source
combination changes. Their joint contribution is evaluated using annual RMSE
and probabilistic skill (Sections~\ref{sec:annual_rmse_results}
and~\ref{sec:probabilistic_results}), while
Section~\ref{sec:mechanism_diagnostics} examines how the likelihood factors
interact during sampling.

\end{enumerate}

\section{Related Work}

Approaches for incorporating observations into atmospheric models differ in whether observation integration is learned during training or specified during inference. We organize the related work into four groups: learned conditioning, inference-time conditioning of pretrained diffusion models, observation-based analysis and forecasting systems, and models operating directly in observation space.

\paragraph{Learned conditioning.}
Learned conditional models incorporate observation information during
training. SLAMS combines gridded in situ and satellite-derived modalities in a
latent score model \citep{qu2024slams}; MODS learns multi-source conditioning
for regional downscaling \citep{tu2025mods}; and PhyDA uses physical
regularization and a reconstruction encoder \citep{wang2025phyda}. 
In these approaches, the relationship between observations and the gridded atmospheric state is learned during training.

\paragraph{Inference-time conditioning.}
Score-based formulations and the Elucidating Diffusion Model (EDM) framework
provide generative priors and reverse-time samplers for inference-time
conditioning \citep{song2021score,karras2022edm}.
Plug-and-play and restoration methods
then reuse a pretrained diffusion model with differentiable constraints or
known physical or linear degradation operators
\citep{song2022medical,graikos2022plugandplay,kawar2022ddrm}. DPS and
pseudoinverse-guided diffusion broaden this strategy to noisy and nonlinear
inverse problems by differentiating measurement consistency through a denoised
estimate \citep{chung2023dps,song2023pigdm}. Tweedie-moment projection improves
this approximation by retaining conditional covariance information
\citep{boys2024tmpd}. 
Sequential Monte Carlo methods \citep{wu2023twisted,dou2024filtering} and Markov chain Monte Carlo (MCMC) methods \citep{wu2024principled} provide alternative routes for sampling a posterior defined by a diffusion prior and likelihood. Recent analyses show that standard DPS may not approximate this posterior faithfully and can produce samples with reduced diversity \citep{xu2025rethinking}. In atmospheric data assimilation, different definitions of the prior and likelihood likewise lead diffusion-based methods to different posterior distributions \citep{hodyss2026diffusionda}.

Atmospheric examples include trajectory assimilation
\citep{rozet2023score}, DiffDA with simulated ERA5 observations
\citep{huang2024diffda}, global latent assimilation with differentiable
operators \citep{andry2025appa}, reverse-process latent optimization
\citep{sun2025losda}, and zero-shot atmospheric inverse problems
\citep{aich2026wind}. Regional studies use real surface stations and evaluate
ensembles against held-out observations \citep{manshausen2025generative,chao2025clin}, while
physics-guided score-based reconstruction has been applied to three-dimensional
tropical-cyclone fields from sparse dropsonde profiles
\citep{han2026tropicalcyclone}. 
Our previous work developed the 13-variable prior and sampler used here and demonstrated conditioning on coarse ERA5 fields, IGRA radiosondes, and atmospheric-emulator outputs without retraining the prior \citep{dj_diffusion_prior}.
Related work extends
inference-time DPS to adaptive sensing and data assimilation in nonlinear
turbulent systems and applies posterior conditioning to diffusion-based climate
downscaling \citep{chakraborty2026adaptive,guan2026climatedownscaling}. 
These studies establish inference-time conditioning as a route
to prior reuse, but do not address how observations from structurally
different in situ sources can be combined around one pretrained atmospheric
prior while keeping their construction, state mapping, residual counting, and
likelihood weighting explicit.

\paragraph{Observation-based analysis and forecasting.}
Some learned systems use observations to produce atmospheric analyses that initialize forecasts or to generate forecasts directly. ADAF combines station and satellite observations with a numerical weather prediction (NWP) background to produce gridded analyses for short-range forecasting \citep{xiang2025adaf}; Aardvark maps observations directly to global gridded and local station forecasts \citep{allen2025aardvark}; and VAE-Var and latent variational assimilation produce analyses by optimizing within learned state representations \citep{xiao2025vaevar,fan2026lda}. These systems learn at least part of the relationship between observations and an analysis or forecast during training, whereas our method combines explicit observation-derived likelihood factors with a separately pretrained atmospheric prior during inference.

\paragraph{Models operating in observation space.}
These models forecast observations directly rather than first reconstructing a regularly gridded atmospheric state. OCELOT separately encodes satellite, radiosonde, aircraft, and surface observations and produces forecasts for the corresponding observing systems \citep{gholoubi2026ocelot}. Our task instead uses observations to reconstruct a common 13-variable gridded atmospheric state at a single analysis time.

\paragraph{Positioning.}
Our method belongs to inference-time conditioning: the pretrained atmospheric prior remains fixed, and each observation source contributes an explicit likelihood factor. The retained measurements, state mapping, residual-counting rule, and likelihood weighting remain explicit for each source. Observation operators, the averaging of densely sampled observations, and error weighting have established precedents in classical data assimilation \citep{carrassi2018dataassimilation}, while mismatches between observation support and model resolution are recognized sources of representation error \citep{janjic2018representation}. Our contribution is not any one of these ingredients in isolation, but their organization as composable observation interfaces that allow structurally different observation sources to constrain the same pretrained atmospheric prior.

\section{Problem Setup}
We first define the pretrained atmospheric prior and one likelihood factor for each observation source, then describe how DPS combines them during reconstruction. All state variables and corresponding observations entering the sampler are standardized using the channel-wise means and standard deviations from ERA5 prior training. Unless otherwise stated, $\mathbf{x}$ and $\mathbf{y}$ below denote quantities in this standardized form.

The channel set is
\begin{equation*}
\mathcal{K}=\{t2m,u10,v10,t500,u500,v500,z500,q500,
t850,u850,v850,z850,q850\}
\end{equation*}
Here $t$, $u$, $v$, and $q$ denote temperature, zonal wind, meridional wind,
and specific humidity, respectively; following ERA5 naming, $z$ denotes
geopotential. Each suffix identifies a surface quantity or pressure level.
Let $\xclean\in\R^{C\times N_\phi\times N_\ell}$ denote a clean atmospheric
state, where $C=|\mathcal{K}|=13$, $N_\phi=128$, and $N_\ell=256$. 

Let $s$ be the fixed standardized land-sea-mask condition supplied to the denoiser, and let $p_\theta(\xclean\mid s)$ denote the atmospheric prior before conditioning on R/A/S observations. This prior is represented by the pretrained 13-variable ERA5 denoiser based
on the EDM framework and developed by \citet{dj_diffusion_prior}.
We use this denoiser and its associated sampler without
retraining the denoiser. The R interface follows the IGRA conditioning design
established in our previous work \citep{dj_diffusion_prior}, while the A and S
interfaces are developed and calibrated here.

In the EDM framework used by the sampler, a noisy
state can be written as
$\xnoisy=\xclean+\sigma\boldsymbol{\epsilon}$ with
$\boldsymbol{\epsilon}\sim\mathcal{N}(0,I)$. The fixed denoiser gives
$\xden=D_\theta(\xnoisy,\sigma,s)$, and the corresponding prior score is
\begin{equation}
    \mathbf{s}_\theta(\xnoisy,\sigma,s)
    =\frac{D_\theta(\xnoisy,\sigma,s)-\xnoisy}{\sigma^2}
    \label{eq:edm_prior_score}
\end{equation}
Equation~\eqref{eq:edm_prior_score} supplies the reverse sampler's prior-score
term.
This clean/noisy/denoised distinction is retained below: observation residuals
are evaluated on a denoised clean-state estimate, whereas likelihood guidance
is differentiated with respect to the current noisy sampler state.

Every state entry $x_{0,kij}$ has a known channel $k$ and coordinate
$(\phi_i,\ell_j)$; each $H_m$ uses location, level, and variable metadata to
map this geolocated state into its observation space.

We now associate the pretrained prior with one observation likelihood factor for each source. Let $\Mset=\{\mathrm{R},\mathrm{A},\mathrm{S}\}$ denote the radiosonde, aircraft, and surface-station observation sources summarized in Figure~\ref{fig:terminology_map}. For source $m\in\Mset$, $\mathbf y_m$ denotes the standardized observation values retained and mapped by that source's interface as described in
Sections~\ref{sec:observation_sources}
and~\ref{sec:observation_factorization}. The parameter $\lambda_m$ weights the contribution of source $m$ to the combined likelihood.

Let $\mathcal{K}_m^{+}$ be the channels with at least one valid target and
$\mathcal{I}_{mk}$ the matched station targets for R or occupied-cell targets
for A and S in active channel $k$. Write $H_{mk}(\xclean)_r$ and $y_{mkr}$ for
the predicted and observed entries indexed by $r$ in active channel $k$.
Define the residual vector and its scalar energy by
\begin{equation}
    \begin{aligned}
    \mathbf{e}_m(\xclean)
    &:=H_m(\xclean)-\mathbf{y}_m,\\
    E_m(\xclean)
    &:=\mathbf{e}_m(\xclean)^\top W_m\mathbf{e}_m(\xclean)\\
    &=\frac{1}{|\mathcal{K}_m^{+}|}
    \sum_{k\in\mathcal{K}_m^{+}}
    \frac{1}{|\mathcal{I}_{mk}|}
    \sum_{r\in\mathcal{I}_{mk}}
    \left(H_{mk}(\xclean)_r-y_{mkr}\right)^2
    \end{aligned}
    \label{eq:modality_loss}
\end{equation}
Thus, each active channel receives equal total weight, and retained
observations or occupied cells are weighted equally within that channel. The
symbol $W_m$ denotes exactly this counting structure.
For one analysis time, let $N_m=\sum_{k\in\mathcal{K}_m^+}
|\mathcal{I}_{mk}|$. After vectorizing the state,
$H_m:\R^{C N_\phi N_\ell}\rightarrow\R^{N_m}$ and
$W_m\in\R^{N_m\times N_m}$. The residual energy measures observation
mismatch; its influence during reverse diffusion is controlled by the
following noise-dependent scale:
\begin{equation}
    v_m(\sigma)
    =
    \mathrm{std}_m^2 + \gamma_m\sigma^2
    \label{eq:variance_schedule}
\end{equation}
where $\mathrm{std}_m$, $\gamma_m$, and $\lambda_m$ are standardized-space
guidance parameters. For a generic
clean-state $\mathbf z$, we represent the likelihood factor for source $m$ using a squared-residual Gaussian form with the noise-dependent variance term $v_m(\sigma)$. 
Up to terms independent of the state,
\begin{equation}
    \log q_{m,\sigma}
    \!\left(\mathbf{y}_m\mid H_m(\mathbf{z})\right)
    =-\frac{E_m(\mathbf{z})}{2v_m(\sigma)}
    +\operatorname{const}(\mathbf{y}_m,\sigma)
    \label{eq:approximate_likelihood_energy}
\end{equation}
The corresponding source-specific negative log-likelihood contributions and their sum are
\begin{equation}
    \mathcal{L}_m(\xclean;\sigma)
    :=\frac{\lambda_m}{2v_m(\sigma)}E_m(\xclean),
    \qquad
    \Lobs(\xclean;\sigma)
    :=\sum_{m\in\Mset}\mathcal{L}_m(\xclean;\sigma)
    \label{eq:additive_loss}
\end{equation}

At each reverse-diffusion noise level, we treat the observation sources as
conditionally independent given the atmospheric state and combine their
likelihood factors in the following clean-state factorization:
\begin{equation}
    \widetilde{p}_\sigma
    \!\left(\xclean\mid\{\mathbf{y}_m\}_{m\in\Mset},s\right)
    \propto
    p_\theta(\xclean\mid s)
    \prod_{m\in\Mset}
    q_{m,\sigma}
    \!\left(\mathbf{y}_m\mid H_m(\xclean)\right)^{\lambda_m}
    \propto p_\theta(\xclean\mid s)
    \exp[-\Lobs(\xclean;\sigma)]
    \label{eq:posterior_factorization}
\end{equation}
Equation~\eqref{eq:posterior_factorization} organizes the observation
constraints in clean-state coordinates. During sampling, DPS evaluates
$\Lobs$ at the denoised estimate $D_\theta(\xnoisy,\sigma,s)$ and
differentiates the resulting composition with respect to $\xnoisy$. This
chain-rule pullback supplies the likelihood correction to the noisy-state
score and is written explicitly in Section~\ref{sec:runtime_guidance}.
Section~\ref{sec:observation_factorization} specifies how each
source constructs $\mathbf y_m$, $H_m$, and the counting
structure inside $E_m$, and how the resulting likelihood contribution is
differentiated through the denoiser to update the noisy sampler state.

Table~\ref{tab:notation} collects the notation used in this setup, including the likelihood-gradient and sampler-update quantities defined in Section~\ref{sec:runtime_guidance}.

\begin{table}[H]
\centering
\small
\renewcommand{\arraystretch}{1.30}
\caption{Core inference and observation-interface notation. Atmospheric states and sampler variables have shape $13\times128\times256$; for source $m$, $\mathbf y_m\in\mathbb R^{N_m}$, $H_m:\mathbb R^{13\cdot128\cdot256}\rightarrow\mathbb R^{N_m}$, and $W_m\in\mathbb R^{N_m\times N_m}$.}

\label{tab:notation}
\begin{tabularx}{\textwidth}{@{}
>{\raggedright\arraybackslash}p{0.22\textwidth}
>{\raggedright\arraybackslash}X@{}}
\toprule
Symbol & Meaning \\
\midrule
$\Mset$, $\mathcal K$ & Observation-source and state-variable index sets. \\
$\mathbf{x}_0$, $\mathbf{x}_\sigma$ & Clean standardized state and its
noise-level-$\sigma$ counterpart. \\
$\mathbf{x}_i$, $\widehat{\mathbf{x}}_{0,i}$ & Current noisy sampler state
and denoised clean-state estimate at reverse step $i$. \\
$\sigma$, $\sigma_i$ & Generic diffusion-noise level and its value at reverse
step $i$. \\
$\mathbf{y}_m$, $H_m$, $W_m$ & Final likelihood targets, observation operator, and residual-counting weights for source $m\in\Mset$. \\
$\lambda_m$, $v_m(\sigma)$ & Likelihood weight and noise-dependent variance term for source $m$. \\
$E_m$, $\mathcal{L}_m$, $\ell_m$ & Weighted squared residual, negative log-likelihood contribution, and
log-likelihood contribution, with $\ell_m=-\mathcal{L}_m$. \\
$\mathbf{d}_m$ & Gradient of the log-likelihood for source $m$ with respect to a denoised clean-state estimate. \\
$\mathbf{g}_{m,i}^{\mathrm{raw}}$ & Runtime guidance after the denoiser-Jacobian
pullback to the noisy sampler state. \\
$\mathbf{c}_i$ & Summed, clipped, and noise-scaled likelihood correction to
the reverse-solver slope. \\
$\Nens$ & Number of conditioned ensemble members; $\Nens=16$ here. \\
\bottomrule
\end{tabularx}
\end{table}

Throughout the paper, \emph{radiosonde (R)}, \emph{aircraft (A)}, and \emph{surface station (S)} denote the three observation sources, and the same labels are used for their corresponding likelihood factors and interfaces.

\section{Observation Sources}
\label{sec:observation_sources}

\begin{figure}[t]
\centering
\includegraphics[width=\textwidth]{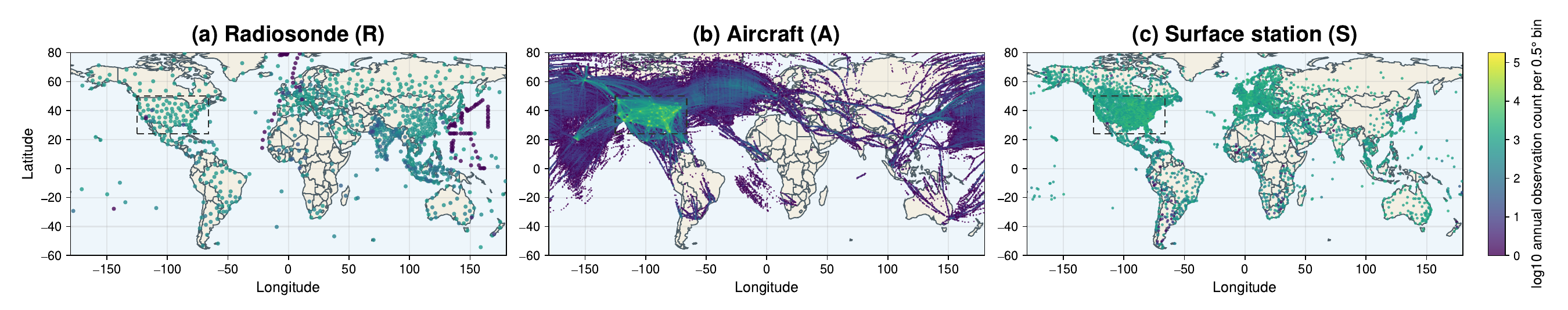}
\caption{Annual spatial distributions of (a) IGRA
radiosonde profiles (R), (b) MADIS ABO aircraft reports (A), and
(c) MADIS METAR surface-station reports (S) during 2020 at 00/12 UTC.
For visualization, observations are grouped into common latitude--longitude
bins of $0.5^{\circ}\times0.5^{\circ}$, and all panels use the same color
scale. For R, each station profile available at an analysis time contributes
one count to its bin; for A and S, each report contributes one count.
Dashed boxes mark the CONUS domain used for A and S.}
\label{fig:observation_geometry}
\end{figure}

We use observations from three products with distinct sampling geometries: sparse IGRA vertical profiles, MADIS ABO aircraft reports concentrated along flight routes, and a dense, irregular network of MADIS METAR surface stations. Figure~\ref{fig:observation_geometry} illustrates these contrasting spatial distributions and the global R and CONUS A/S domains used in this study. We note that while the three observation sources have different sampling densities, their influence on the likelihood is determined by source-specific residual counting and likelihood parameters rather than by report count alone. For A and S, reports are averaged within each model-grid cell separately for each variable, so each occupied cell contributes one residual for that variable. 
Table~\ref{tab:modalities} summarizes the sampling geometry,
variables, and spatial domain of each source.

\begin{table}[t]
\centering
\small
\renewcommand{\arraystretch}{1.24}
\caption{Observation sources, sampling geometry, target variables, and spatial domains used in this study.}
\label{tab:modalities}
\begin{tabular}{p{0.14\textwidth}p{0.30\textwidth}p{0.24\textwidth}p{0.20\textwidth}}
\toprule
Source & Product and sampling geometry & Variables & Spatial domain\\
\midrule
Radiosonde (R) & IGRA sparse vertical profiles & $t2m$, $u10$, $v10$, $t500$, $u500$, $v500$, $z500$, $q500$, $t850$, $u850$, $v850$, $z850$, $q850$ & Global \\
\specialrule{0.3pt}{2.5pt}{2.5pt}
Aircraft (A) & MADIS ABO reports clustered along flight routes & $t500$, $u500$, $v500$, $t850$, $u850$, $v850$ & CONUS domain \\
\specialrule{0.3pt}{2.5pt}{2.5pt}
Surface station (S) & MADIS METAR surface stations &
$t2m$, $u10$, $v10$ & CONUS domain \\
\bottomrule
\end{tabular}
\end{table}

\FloatBarrier
\paragraph{Radiosonde observations (R).}

IGRA provides vertical profiles of temperature, geopotential height, humidity,
and horizontal wind from a global station network, with pressure identifying
the vertical location of each measurement
\citep{noaa_ncei_igra,durre2016igrav2,durre2018igra}. Following \citet{dj_diffusion_prior}, IGRA geopotential height at 500 and
850~hPa is converted to geopotential before standardization. The R interface
provides global conditioning by associating the IGRA surface-temperature and
surface-wind fields with $t2m$, $u10$, and $v10$, and observations reported at
500 and 850~hPa with the corresponding pressure-level state channels. The
R-only configuration serves as the reference for evaluating the additions of
A and S.

\paragraph{Aircraft observations (A).}
The MADIS ABO product contains temperature and wind measurements from several aircraft reporting systems \citep{miller2005madis,noaa_madis_aircraft}. We use individual reports sampled along flight trajectories. Because these reporting systems differ in coverage and observation quality \citep{wagner2021airborne}, we compared candidate combinations during interface development. The final A interface uses ACARS direct, MDCRS/ARINC, and Canadian AMDAR reports \citep{noaa_madis_upperair_notes}. Streams with no or only sparse eligible observations were excluded, while TAMDAR was excluded because its inclusion increased reconstruction errors. Appendix~\ref{sec:calibration_inventory} reports this comparison and the resulting selection. 
MADIS provides a quality-control assessment for each reported field. We retain aircraft reports only when the altitude and relevant temperature or wind measurements pass these checks \citep{noaa_madis_aircraft_qc}. We additionally require valid coordinates and finite values and restrict temperature to $180<T<330$ K and wind speed to $0\leq V<150$ m~s$^{-1}$.
Because the atmospheric state represents temperature and horizontal winds at 500 and 850 hPa, we convert reported aircraft altitude to pressure using a standard atmosphere \citep{usstandardatmosphere1976} and assign reports within $\pm25$ hPa of either level to the corresponding model channel.

\paragraph{Surface-station observations (S).}
The MADIS METAR product provides surface temperature and wind observations from a global station network \citep{noaa_madis_surface}. We use stations within the CONUS domain to constrain $t2m$, $u10$, and $v10$. For each 00/12 UTC analysis time, we consider reports within $\pm60$ minutes. We retain only temperature and wind measurements that pass the corresponding MADIS quality-control checks and satisfy finite-value and physical-range criteria \citep{noaa_madis_surface_qc}. Temperature is restricted to $180<T<330$ K and wind speed to $0\leq V<75$ m~s$^{-1}$. The retained reports are then averaged within model-grid cells as described in Section~\ref{sec:surface_interface}.

\section{Composable Observation-Interface Factorization}
\label{sec:observation_factorization}

The retained observations described above define the final
likelihood targets $\mathbf y_m$ for each source. Their different sampling
geometries require different state mappings and residual-counting rules:
radiosondes retain measurements at sparse station locations, whereas aircraft
and surface-station reports are aggregated on the model grid so that report
multiplicity alone does not determine likelihood influence. Each observation
interface is specified by four components:
\begin{enumerate}
    \item \textbf{retained-observation construction:} source
    selection, quality control, temporal and vertical matching, and aggregation
    define $\mathbf y_m$;
    \item \textbf{state mapping:} $H_m$ specifies the
    locations, levels, and state variables at which those targets constrain the
    gridded state;
    \item \textbf{residual counting:} $W_m$ specifies how
    retained station targets or occupied-cell targets and active channels
    contribute to the residual energy; and
    \item \textbf{likelihood weighting:}
    $\lambda_m/v_m(\sigma)$ specifies the calibrated, noise-dependent
    contribution of the source during reverse diffusion.
\end{enumerate}
Together, $\mathbf y_m$, $H_m$, $W_m$, and
$\lambda_m/v_m(\sigma)$ specify which observations enter the likelihood, where
they constrain the state, how their residuals are counted, and how strongly
they influence sampling. Equation~\eqref{eq:modality_loss} assigns equal total
weight to each active channel and equal weight to the retained targets within
that channel. R retains station-level targets evaluated by interpolation,
whereas A and S use occupied-cell targets obtained by within-cell averaging.

\subsection{Radiosonde Interface (R)}

The R operator selects the state channel corresponding to the
observed variable and pressure level and evaluates that channel at each
radiosonde location using bilinear interpolation:
\begin{equation}
    \Higra(\xclean)_r
    =
    \sum_{j\in\mathcal{N}(\mathrm{lat}_r,\mathrm{lon}_r)}
    w_{rj}\,x_{0,k(r),j},
    \label{eq:igra_operator}
\end{equation}
Here, $k(r)$ is the corresponding variable and pressure-level
channel, $(\mathrm{lat}_r,\mathrm{lon}_r)$ is the station location,
$\mathcal{N}(\mathrm{lat}_r,\mathrm{lon}_r)$ contains the four neighboring
model-grid points, and $w_{rj}$ are the bilinear interpolation weights. Because
R observations are spatially sparse, retaining one residual for each available
station target preserves its location rather than first averaging targets on
the model grid.

\subsection{Equal-Cell Interfaces for Aircraft and Surface Stations (A/S)}
\label{sec:equal_cell_interfaces}
\label{sec:surface_interface}

Aircraft reports cluster along flight routes and around
airports, and multiple surface-station reports can also fall within one
model-grid cell. Counting every report as a separate residual can therefore
give densely sampled locations disproportionate likelihood influence and can
amplify correlations associated with unresolved spatial support
\citep{janjic2018representation}. For A and S, we instead average reports
within model-grid cells and retain one residual for each occupied cell and
observed variable. For $m\in\{\mathrm{A},\mathrm{S}\}$,
\begin{equation}
    \bar{y}_{m,k,a}
    =
    \frac{1}{|\mathcal{P}_{m,k,a}|}
    \sum_{p \in \mathcal{P}_{m,k,a}} y_{m,k,p},
    \label{eq:cell_mean}
\end{equation}
where $\mathcal{P}_{m,k,a}$ is the set of standardized
reports from source $m$ and channel $k$ assigned to the cell associated with
the nearest model-grid point $a$. Let $\mathcal{A}_{mk}$ denote the occupied
cells for that source and channel. For A and S, $\mathbf y_m$ consists of these
cell means, and $H_m$ selects the corresponding state values at the associated
grid points. The residual energy first averages over occupied cells within
each active channel and then averages over the active channels:
\begin{equation}
    E_m(\xclean)
    =
    \frac{1}{|\mathcal{K}_m^{+}|}
    \sum_{k\in\mathcal{K}_m^{+}}
    \frac{1}{|\mathcal{A}_{mk}|}
    \sum_{a\in\mathcal{A}_{mk}}
    (x_{0,k,a}-\bar{y}_{m,k,a})^2.
    \label{eq:equal_cell_loss}
\end{equation}

The A interface applies this construction to temperature and
horizontal winds at 500 and 850 hPa, while the S interface applies it to
$t2m$, $u10$, and $v10$. A and S therefore directly constrain disjoint sets of
state variables, while R overlaps both sets and can also constrain the
remaining state variables.

\begin{figure}[H]
    \centering
    \includegraphics[width=0.98\textwidth]{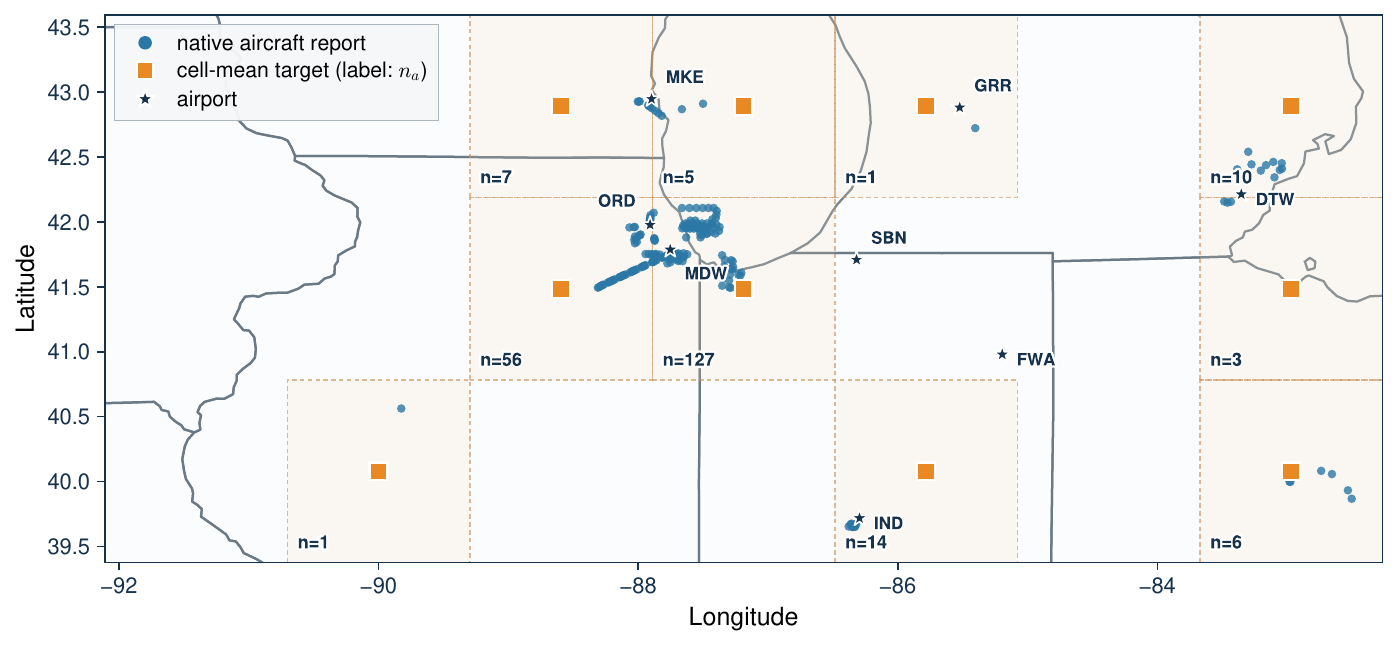}
    \caption{Equal-cell construction illustrated with
    retained 850-hPa aircraft temperature reports near Chicago. Blue points
    show individual reports, shaded rectangles show occupied model-grid cells,
    and orange squares show the associated grid points. The label $n_a$ gives
    the number of reports averaged to form the cell-mean target $\bar y_a$.
    Stars mark airport locations. Each occupied cell contributes
    one residual regardless of the number of reports it contains.}
    \label{fig:equal_cell_operator}
\end{figure}

Figure~\ref{fig:equal_cell_operator} illustrates the
distinction between report density and residual counting using aircraft
observations. Multiple reports can refine a cell-mean target, but the occupied
cell still contributes one residual for that variable.

\subsection{Inference-Time Composition of Observation Factors}

Once $\mathbf y_m$, $H_m$, $W_m$, and the likelihood
parameters are specified, each source contributes one additive negative
log-likelihood term $\mathcal L_m$. For R, A, and S,
Equation~\eqref{eq:additive_loss} becomes
\begin{align}
    \Lobs(\xclean;\sigma)
    &=\sum_{m\in\Mset}
    \frac{\lambda_m E_m(\xclean)}{2v_m(\sigma)} \notag\\
    &=\frac{\lambda_{\mathrm{R}}E_{\mathrm{R}}(\xclean)}
    {2v_{\mathrm{R}}(\sigma)}
    +\frac{\lambda_{\mathrm{A}}E_{\mathrm{A}}(\xclean)}
    {2v_{\mathrm{A}}(\sigma)}
    +\frac{\lambda_{\mathrm{S}}E_{\mathrm{S}}(\xclean)}
    {2v_{\mathrm{S}}(\sigma)}.
    \label{eq:tri_loss}
\end{align}

The R likelihood parameters are inherited from the previous
study, whereas the A and S parameters are selected as described in
Section~\ref{sec:independent_year_development}. Table~\ref{tab:final_protocols}
reports the values used in the evaluation.

\subsection{From Observation Residuals to Runtime DPS Guidance}
\label{sec:runtime_guidance}

At each reverse step that updates the sampler state, DPS
evaluates the observation mismatch using a denoised estimate and
differentiates that mismatch with respect to the current noisy state through
the denoiser \citep{chung2023dps}. Let $i$ index the reverse steps,
$\mathbf{x}_i$ denote the current noisy state, and $\sigma_i$ denote its noise
level. The denoiser produces
\begin{equation}
    \widehat{\mathbf{x}}_{0,i}
    =D_\theta(\mathbf{x}_i,\sigma_i,s),
    \label{eq:denoised_estimate}
\end{equation}
which estimates the clean standardized atmospheric state.
For a candidate clean state $\mathbf z$, write the weighted log-likelihood
contribution from source $m$ as
$\ell_m(\mathbf z;\sigma):=-\mathcal L_m(\mathbf z;\sigma)$.
We call its gradient with respect to the candidate clean
state the \emph{direct likelihood gradient}:
\begin{align}
    \mathbf{d}_m(\mathbf{z};\sigma)
    &:=\nabla_{\mathbf{z}}\ell_m(\mathbf{z};\sigma)
    =-\nabla_{\mathbf{z}}\mathcal{L}_m(\mathbf{z};\sigma) \notag\\
    &=-\frac{\lambda_m}{v_m(\sigma)}
    J_{H_m}(\mathbf{z})^\top W_m
    \left[H_m(\mathbf{z})-\mathbf{y}_m\right],
    \label{eq:direct_modality_guidance}
\end{align}
Here, $J_{H_m}$ is the Jacobian of the state mapping. For the
bilinear interpolation used by R and the grid-point selection used by A and S,
it is the corresponding interpolation or selection matrix.
Equation~\eqref{eq:direct_modality_guidance} separates the roles of the
interface components. The residual measures disagreement with the retained
targets; $W_m$ assigns equal total weight to each active channel and equal
weight to the retained station targets for R or occupied-cell targets for A
and S; $\lambda_m/v_m(\sigma)$ controls the noise-dependent likelihood
weighting; and $J_{H_m}^{\top}$ maps the residual correction to the
corresponding state variables and locations.

The sampler does not update
$\widehat{\mathbf{x}}_{0,i}$ directly. Instead, differentiation through the
denoiser maps the likelihood gradient to the current noisy state. Define
\begin{align}
    J_{D,i}
    &:=\frac{\partial
    D_\theta(\mathbf{x}_i,\sigma_i,s)}
    {\partial\mathbf{x}_i}, \notag\\
    \mathbf{g}_{m,i}^{\mathrm{raw}}
    &:=\nabla_{\mathbf{x}_i}
    \ell_m\!\left(
    D_\theta(\mathbf{x}_i,\sigma_i,s);
    \sigma_i\right)
    =J_{D,i}^\top
    \mathbf{d}_m(\widehat{\mathbf{x}}_{0,i};\sigma_i).
    \label{eq:runtime_modality_guidance}
\end{align}
Thus, $H_m$ determines the locations and variables on which a
source acts directly, while $J_{D,i}^{\top}$ couples that information across
the gridded multivariate state through the pretrained denoiser. Let
$\mathbf{s}_\theta(\mathbf{x}_i,\sigma_i,s)$ denote the prior score implied by
the denoiser, and let $\mathbf{s}_{\mathrm{post},i}$ denote the approximate
posterior score with respect to the current noisy state. Following DPS
\citep{chung2023dps}, the score before clipping is
\begin{equation}
    \mathbf{s}_{\mathrm{post},i}
    \approx
    \mathbf{s}_\theta
    (\mathbf{x}_i,\sigma_i,s)
    +\sum_{m\in\Mset}\mathbf{g}_{m,i}^{\mathrm{raw}},
    \label{eq:posterior_score_decomposition}
\end{equation}
where the likelihood gradients are evaluated through the
denoised estimate. Because differentiation is linear, the gradient of the
summed log-likelihood equals the sum of the source-specific gradients. The
combined gradient is clipped once elementwise and then scaled by $\sigma_i$:
\begin{equation}
    \mathbf{c}_i
    =\sigma_i\,
    \operatorname{clip}_{[-1,1]}\!\left(
    \sum_{m\in\Mset}\mathbf{g}_{m,i}^{\mathrm{raw}}\right).
    \label{eq:runtime_applied_guidance}
\end{equation}
In the EDM parameterization, the current prior-derived
reverse-solver slope is
$(\mathbf{x}_i-\widehat{\mathbf{x}}_{0,i})/\sigma_i
=-\sigma_i\mathbf{s}_\theta(\mathbf{x}_i,\sigma_i,s)$
\citep{karras2022edm}. The sampler subtracts $\mathbf c_i$ from the
Heun-averaged prior-derived slope, and the reverse-step size converts the
resulting slope into a state update. Source-specific gradients shown in later
diagnostics represent the components before the shared sum and clipping, so
they need not sum to the applied correction after clipping.

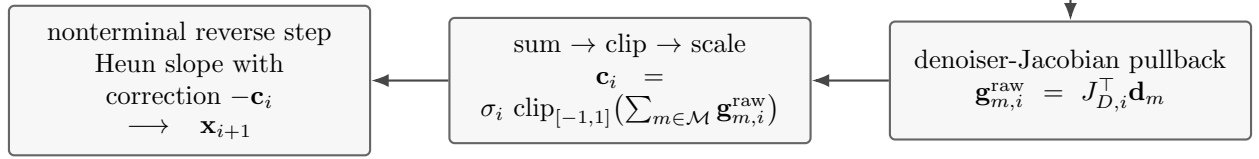
\begin{figure}[H]
\centering
\begin{tikzpicture}[
    font=\small,
    >=Latex,
    state/.style={draw=#1, very thick, rounded corners=2pt, fill=#1!8,
        align=center, minimum height=1.18cm, text width=4.00cm, inner sep=6pt},
    modality/.style={draw=#1, thick, rounded corners=2pt, fill=#1!7,
        align=center, minimum height=2.42cm, text width=4.45cm, inner sep=8pt},
    process/.style={draw=black!60, thick, rounded corners=2pt, fill=black!3,
        align=center, minimum height=1.56cm, text width=4.35cm, inner sep=6pt},
    common/.style={draw=black!60, thick, rounded corners=2pt, fill=black!3,
        align=center, minimum height=1.42cm, text width=9.2cm, inner sep=7pt},
    flow/.style={->, thick, color=black!72},
    bus/.style={thick, color=black!62}
]
\definecolor{priorblue}{HTML}{2769A5}
\definecolor{igrablue}{HTML}{2563A6}
\definecolor{aboorange}{HTML}{D97706}
\definecolor{metarteal}{HTML}{00897B}

\node[anchor=west, font=\bfseries\small] at (-8.15,8.15)
    {1. Estimate the clean atmospheric state};
\node[state=priorblue] (xs) at (-6.00,7.05)
    {current noisy state\\$\mathbf{x}_i$};
\node[state=priorblue] (denoiser) at (0,7.05)
    {fixed denoiser\\$D_\theta(\mathbf{x}_i,\sigma_i,s)$};
\node[state=priorblue] (xzero) at (6.00,7.05)
    {denoised atmospheric state\\$\widehat{\mathbf{x}}_{0,i}\in\R^{13\times128\times256}$};

\node[anchor=west, font=\bfseries\small] at (-8.15,5.92)
    {2. Apply a geometry-specific observation interface};
\node[modality=igrablue] (igra) at (-5.70,3.30)
    {\textbf{Radiosonde (R)}\\[2pt]
    variable and level matching\\bilinear interpolation $H_{\rm R}$\\equal target and channel weights $W_{\rm R}$};
\node[modality=aboorange] (abo) at (0,3.30)
    {\textbf{Aircraft (A)}\\[2pt]
    source selection and pressure matching\\cell means and grid-point selection $H_{\rm A}$\\
    equal cell and channel weights $W_{\rm A}$};
\node[modality=metarteal] (metar) at (5.70,3.30)
    {\textbf{Surface station (S)}\\[2pt]
    time-matched reports and cell means\\grid-point selection $H_{\rm S}$\\
    equal cell and channel weights $W_{\rm S}$};

\node[common] (direct) at (0,0.35)
    {\textbf{Shared residual formulation, evaluated separately for each source $m$}\\
    $\mathbf e_m=H_m(\widehat{\mathbf{x}}_{0,i})-\mathbf y_m$\qquad
    $\mathbf d_m=\nabla_{\widehat{\mathbf{x}}_{0,i}}
    \ell_m(\widehat{\mathbf{x}}_{0,i};\sigma_i)$};

\node[anchor=west, font=\bfseries\small] at (-8.15,-0.76)
    {3. Pull back, combine, and update};
\node[process] (pullback) at (5.82,-2.10)
    {denoiser-Jacobian pullback\\$\mathbf g_{m,i}^{\rm raw}=J_{D,i}^{\top}\mathbf d_m$};
\node[process] (combine) at (0,-2.10)
    {sum $\rightarrow$ clip $\rightarrow$ scale\\
    $\mathbf c_i=\sigma_i\,
    \operatorname{clip}_{[-1,1]}\!\left(\sum_{m\in\Mset}
    \mathbf g_{m,i}^{\rm raw}\right)$};
\node[process] (update) at (-5.82,-2.10)
    {nonterminal reverse step\\
    Heun slope with correction $-\mathbf c_i$\\[-1pt]
    $\longrightarrow\ \mathbf{x}_{i+1}$};

\draw[flow] (xs) -- (denoiser);
\draw[flow] (denoiser) -- (xzero);
\coordinate (operatorright) at (5.70,5.32);
\coordinate (operatorcenter) at (0,5.32);
\coordinate (operatorleft) at (-5.70,5.32);
\coordinate (xzerobus) at (6.00,5.32);
\draw[bus] (xzero.south) -- (xzerobus) -- (operatorleft);
\draw[flow] (operatorleft) -- (igra.north);
\draw[flow] (operatorcenter) -- (abo.north);
\draw[flow] (operatorright) -- (metar.north);
\draw[flow] (igra.south) -- (direct.north west);
\draw[flow] (abo.south) -- (direct.north);
\draw[flow] (metar.south) -- (direct.north east);
\draw[flow] (direct.south east) -- ++(0,-0.20) -| (pullback.north);
\draw[flow] (pullback) -- (combine);
\draw[flow] (combine) -- (update);
\end{tikzpicture}
\caption{DPS conditioning with composable observation
interfaces. Each source defines its retained targets, state mapping, and
residual-counting rule. The resulting likelihood gradients are mapped through
the shared denoiser, combined, clipped, and applied as a correction to the
reverse-solver update.}
\label{fig:dps_mechanism_flow}
\end{figure}

Figure~\ref{fig:dps_mechanism_flow} summarizes the sequence
from source-specific observation residuals through the shared denoiser
pullback to the combined likelihood correction and reverse-solver update.

\FloatBarrier

\section{Evaluation Design and Metrics}
\label{sec:evaluation_design}

The evaluation is designed to isolate the contributions of A
and S while preventing the 2020 results from influencing interface
development.
We use observations from 2019 to select the A and S interface
designs and likelihood parameters, then evaluate the selected interfaces
independently throughout 2020 without further tuning.
The pretrained diffusion prior is not trained or fine-tuned
using either the 2019 or 2020 cases.
Conditioning comparisons are matched so that only the active
observation likelihood factors differ. We evaluate these comparisons using
reconstruction errors on the model grid and at held-out observations,
probabilistic metrics, and uncertainty intervals that account for temporal
dependence.

\subsection{Fixed Prior, Matched Sampling, and Evaluation Domains}
\label{sec:evaluation_domains}

At each analysis time, every configuration uses the same
13-variable atmospheric prior and sampler to generate a 16-member ensemble of
stochastic reconstructions, with 50 denoising steps and matched member seeds. Appendix~\ref{sec:run_lineage} reports the
additional sampling parameters and the analysis times used
for the selected single-case figures. ERA5 fields on the same model grid provide the
reference states used to calculate gridded reconstruction errors
\citep{hersbach2020era5}. We report the
gridded results over three domains:
\begin{itemize}
    \item the CONUS domain, containing model-grid points
    whose centers lie within $24$--$50^{\circ}$N and
    $125$--$66^{\circ}$W;
    \item the global domain; and
    \item the global domain excluding CONUS, used to measure
    aggregate changes outside the region directly constrained by A and S.
\end{itemize}
As defined in the Introduction, the CONUS domain is this fixed
latitude--longitude rectangle rather than a political-boundary or land-only
mask. It contains 19 latitude rows and 42 longitude columns, giving 798 model-grid
points. Aircraft and surface-station reports are retained when their native
coordinates lie within the same rectangle and are subsequently assigned to
the cells associated with their nearest model-grid points. The gridded
evaluation mask is defined from model-grid-point centers and does not depend
on the number or locations of the retained reports.

\subsection{Interface Development and Independent-Year Evaluation}
\label{sec:independent_year_development}

All interface comparisons and likelihood-parameter selection
use 24 development cases from 2019 at 00 UTC on the 1st and
15th of every month. R provides the common reference throughout development. 
The three stages select and calibrate the A and S interfaces and 
then determine how they are combined with R:

\begin{enumerate}

    \item select the A interface structure by comparing
    combinations of aircraft reporting streams,
    pressure-matching windows, and pointwise or cell-based residual
    representations; select the S interface structure by comparing its candidate residual representations;

    \item calibrate the likelihood parameters separately for
    the selected A and S interfaces using the R+A and R+S configurations,
    respectively; and

    \item form nine R+A+S parameter settings from candidate
    likelihood-parameter values for A and S and select the final setting for
    the independent 2020 evaluation.

\end{enumerate}

These comparisons select the ${\pm}25$-hPa
pressure windows and equal-cell
residual counting for A, equal-cell
residual counting for S, and the ACARS
direct, MDCRS/ARINC, and Canadian AMDAR streams described above.
Additional non-TAMDAR streams supplied no or only sparse
eligible observations and did not materially change the 2019 reconstruction
errors, whereas including TAMDAR increased them.

In the third stage, a paired-month bootstrap compares each
R+A+S candidate with the candidate producing the largest mean all-variable
RMSE reduction relative to R-only conditioning. Each bootstrap replicate
resamples the 12 calendar months with replacement while retaining the two
development cases within each selected month. We use 10{,}000 replicates to
estimate a 95\% interval for the paired difference in RMSE change. Candidates
whose intervals include zero are treated as indistinguishable under this
criterion. Among these candidates, we select the setting closest to the center
of the tested parameter range, using the lower combined likelihood weight as a
final tie-breaker. This procedure avoids selecting an unnecessarily extreme
setting based only on a small RMSE difference that is not stable across the
2019 months.

The independent 2020 evaluation contains 723 of the 732 nominal 00/12 UTC
analysis times in 2020, namely those for which all inputs
required for R-only conditioning are available.
No 2020 observation or reconstruction result is used to
select or recalibrate the interfaces.
Appendix~\ref{sec:calibration_inventory} reports the complete
development comparisons and selected values.

\subsection{Selected Interfaces and Conditioning Comparisons}

We evaluate the selected interfaces in four matched
conditioning configurations: R-only, R+A, R+S, and R+A+S. All four use the same
prior, sampler settings, and random seeds. R+A and R+S isolate the respective
additions of A and S to the R-only reference, while R+A+S evaluates their combined
contribution. Table~\ref{tab:final_protocols} records the interface designs and likelihood parameters used in these comparisons.

\begin{table}[t]
\centering
\caption{Observation-interface designs and likelihood parameters used in the
2020 evaluation. The R settings are inherited from the previous study, while
the A and S settings are selected using the 2019 development cases.}
\label{tab:final_protocols}
\footnotesize
\setlength{\tabcolsep}{2.5pt}
\renewcommand{\arraystretch}{1.35}
\begin{tabular}{@{}
  >{\raggedright\arraybackslash}p{0.115\textwidth}
  >{\raggedright\arraybackslash}p{0.350\textwidth}
  >{\raggedright\arraybackslash}p{0.140\textwidth}
  >{\raggedright\arraybackslash}p{0.160\textwidth}
  >{\raggedright\arraybackslash}p{0.190\textwidth}@{}}
\toprule
Source & Retained observations &
Constrained variables &
State mapping and residual counting &
Likelihood parameters \\
\midrule

Radiosonde\newline (R) &
IGRA profiles &
$t2m$, $u10$, $v10$,\newline
$t500$, $u500$, $v500$,\newline
$z500$, $q500$,\newline
$t850$, $u850$, $v850$,\newline
$z850$, $q850$ &
bilinear interpolation; equal target and channel weights &
$\lambda_{\mathrm R}=1.0$\newline
$\mathrm{std}_{\mathrm R}=5\times10^{-4}$\newline
$\gamma_{\mathrm R}=2\times10^{-6}$ \\

\specialrule{0.3pt}{2.5pt}{2.5pt}

Aircraft\newline (A) &
MADIS ABO reports from ACARS direct, MDCRS/ARINC, and Canadian AMDAR;
${\pm}25$-hPa windows &
$t500$, $u500$, $v500$,\newline
$t850$, $u850$, $v850$ &
cell means; equal cell and channel weights &
$\lambda_{\mathrm A}=0.4$\newline
$\mathrm{std}_{\mathrm A}=5\times10^{-4}$\newline
$\gamma_{\mathrm A}=2\times10^{-5}$ \\

\specialrule{0.3pt}{2.5pt}{2.5pt}

Surface station\newline (S) &
MADIS METAR reports &
$t2m$, $u10$, $v10$ &
cell means; equal cell and channel weights &
$\lambda_{\mathrm S}=0.4$\newline
$\mathrm{std}_{\mathrm S}=1.25\times10^{-4}$\newline
$\gamma_{\mathrm S}=4\times10^{-5}$ \\

\bottomrule
\end{tabular}
\end{table}

\subsection{Held-Out Observation Evaluation}
\label{sec:heldout_design}

To complement the gridded comparison with ERA5, separate
held-out evaluations assess whether A and S improve predictions at observations
that are not used for conditioning. These evaluations use 24 seasonally
distributed 2020 cases and random splits selected before examining the
results.
For each case, observations from the evaluated source are first restricted to
the CONUS domain using their native reported coordinates. We then exclude
20\% of the surface-station cells or aircraft cell--variable targets from the corresponding
likelihood factor. The remaining 80\% are used for
conditioning, while the excluded observations are used only
for evaluation.

For the S evaluation, R+A+S conditioned on the retained S
cells is compared with R+A at the excluded S cells. For the A evaluation,
R+A+S conditioned on the retained A targets is compared with R+S at the
excluded A cell--variable targets.  For each variable and
analysis time, we calculate the percentage change in RMSE at
the excluded targets and then average with equal weight
across variables and analysis times. 
Uncertainty is estimated using 10{,}000 paired-analysis-time
bootstrap replicates, each obtained by resampling the 24 analysis times with
replacement while retaining the paired errors from the two compared
configurations. Each configuration used in these evaluations
has a 16-member ensemble, with the same member seeds used for the two
configurations in each comparison.

\subsection{Metrics}
For conditioning configuration $P$, let
$\operatorname{RMSE}_{\tau,k,G}^{P}$ denote the ensemble-mean RMSE at analysis
time $\tau$ for state variable $k$ over evaluation domain $G$. Our primary
deterministic metric is its percentage change relative to R-only conditioning
at the matching $\tau$, $k$, and $G$:
\begin{equation}
    \Delta_{\tau,k,G}^{P}
    =
    100
    \frac{\operatorname{RMSE}_{\tau,k,G}^{P}
    -\operatorname{RMSE}_{\tau,k,G}^{\mathrm{R}}}
    {\operatorname{RMSE}_{\tau,k,G}^{\mathrm{R}}}.
    \label{eq:percent_delta}
\end{equation}
The matching indices make
$\Delta_{\tau,k,G}^{P}$ a paired percentage change in RMSE. In the gridded
evaluation, we use \emph{RMSE change} as shorthand for this quantity. Negative
values indicate lower RMSE for configuration $P$ than for R-only conditioning. We
summarize three variable groups:
\begin{itemize}
    \item \emph{all 13 variables}: the complete modeled state;
    \item \emph{surface-targeted variables}: $t2m$, $u10$, and $v10$;
    \item \emph{upper-air temperature and wind variables}: $t500$, $u500$,
    $v500$, $t850$, $u850$, and $v850$.
\end{itemize}
The final group contains the temperature and horizontal-wind
variables directly constrained by A. For a variable set
$\mathcal K'\subseteq\mathcal K$ and the matched set of evaluation times
$\mathcal T$, the grouped mean RMSE change is
\begin{equation}
    \overline{\Delta}_{G,\mathcal{K}'}^{P}
    =\frac{1}{|\mathcal{T}|\,|\mathcal{K}'|}
    \sum_{\tau\in\mathcal{T}}\sum_{k\in\mathcal{K}'}
    \Delta_{\tau,k,G}^{P}.
    \label{eq:headline_mean_of_ratios}
\end{equation}
This dimensionless average gives equal weight to every
analysis time and variable, allowing variables with different physical units
to contribute equally. For a single variable, the per-variable mean RMSE
change is $\Delta_{k,G}^{P}:=\overline{\Delta}_{G,\{k\}}^{P}$. The grouped
and per-variable results therefore use the same aggregation. The corresponding
annual-mean RMSE values in physical units are reported separately in
Table~\ref{tab:absolute_rmse_full723}. Appendix
Section~\ref{sec:metric_definitions} gives the complete metric definitions.

Because RMSE evaluates the ensemble mean rather than the full
ensemble distribution, we separately assess probabilistic skill and
dispersion following standard weather-verification practice
\citep{hersbach2000crps,garg2022weatherbenchprobability,price2025gencast,
manshausen2025generative}.
CRPS is a proper score that rewards both accuracy
and a useful ensemble distribution, with lower values preferred
\citep{gneiting2007scoring}. Spread--skill ratios, rank histograms, and
5th--95th percentile coverage then diagnose ensemble dispersion and
calibration \citep{fortin2014spread,hamill2001rank}. Their finite-ensemble definitions and interpretation are given in
Appendix Sections~\ref{sec:metric_definitions} and
\ref{sec:finite_ensemble_coverage}.

\subsection{Temporal Uncertainty under Serial Correlation}

For the full-year 2020 evaluation, successive 12-hour analysis
times can belong to the same weather system, so their paired reconstruction
errors need not be temporally independent. The primary uncertainty analysis
therefore uses a moving-block bootstrap to preserve temporal dependence within
sampled blocks \citep{kunsch1989bootstrap}.
Following the circular-block construction of
\citet{shao1993circular},
we use 14-day blocks while preserving genuine missing-data
gaps in the original 00/12 UTC calendar.
Within each replicate, the RMSE changes for all variables at a selected
analysis time are retained together, and the grouped mean RMSE change
$\overline{\Delta}_{G,\mathcal K'}^{P}$ defined in
Equation~\eqref{eq:headline_mean_of_ratios} is recomputed.
We report 2.5th--97.5th percentile intervals from
$B=10{,}000$ replicates.
Appendix Section~\ref{sec:supplemental_annual} gives additional details for
the per-variable intervals from the 14-day moving-block bootstrap.
These intervals assess whether the reported annual differences
persist after accounting for temporal dependence, completing the evaluation
design used in the following Results section.

\section{Results}

We evaluate the R-only, R+A, R+S, and R+A+S conditioning
configurations over the independent 2020 evaluation period. We first quantify
annual RMSE changes from adding A and S to R-only conditioning.
We then assess their robustness to temporal dependence, evaluate predictions at
observations excluded from conditioning, and determine whether the
improvements extend from ensemble-mean RMSE to probabilistic skill.

\subsection{Annual RMSE Changes from Composed Observation Sources}
\label{sec:annual_rmse_results}

\begin{figure}[p]
    \centering
    \includegraphics[width=0.75\textwidth]
    {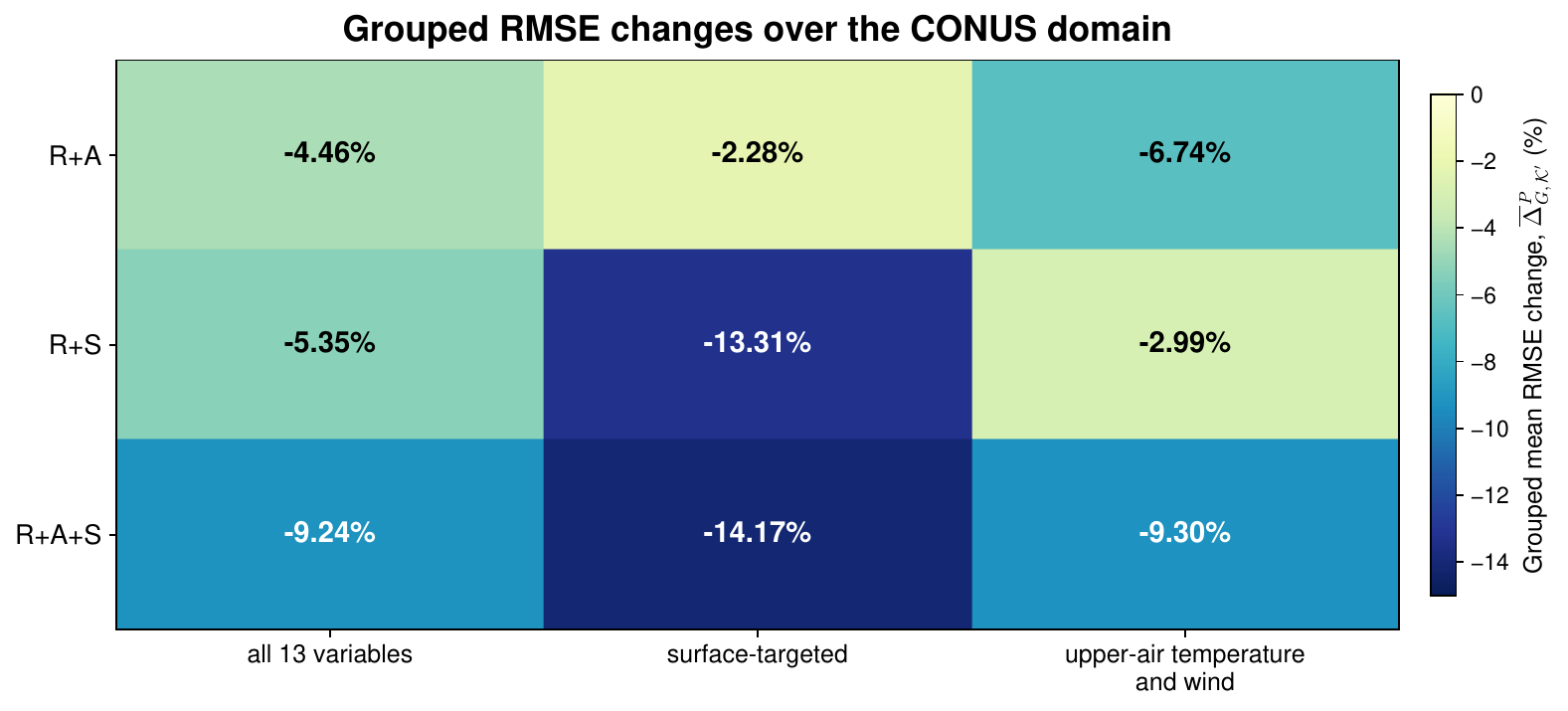}
    \caption{Grouped mean RMSE changes
    $\overline{\Delta}_{G,\mathcal K'}^{P}$ over the
    CONUS domain for R+A, R+S, and R+A+S relative to R-only conditioning across
    723 matched analysis times. Negative values indicate lower RMSE.}
    \label{fig:frozen2019_composition_summary}
\end{figure}

\begin{figure}[p]
    \centering
    \includegraphics[width=1.0\textwidth]
    {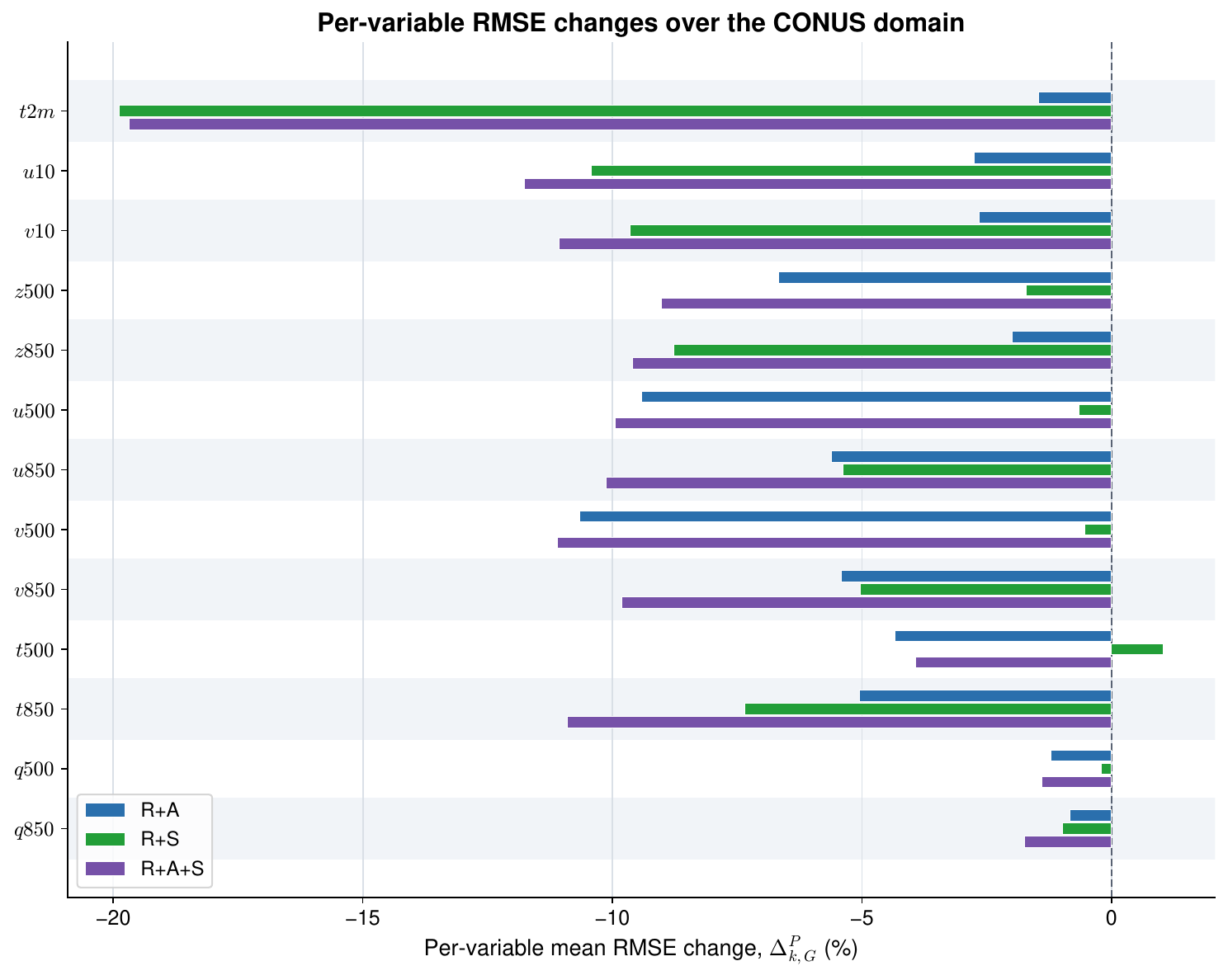}
    \caption{Per-variable mean RMSE changes
    $\Delta_{k,G}^{P}$ for R+A, R+S, and R+A+S over the CONUS domain. Values are
    averaged over 723 matched 2020 analysis times. Negative values indicate
    lower RMSE relative to R-only conditioning.}
    \label{fig:frozen2019_per_variable}
\end{figure}

Table~\ref{tab:full_year_main} reports the primary evaluation over all 723
matched 2020 analysis times. 
Over the CONUS domain, R+A lowers RMSE evaluated against ERA5 for the
upper-air temperature and wind group by $6.74\%$ relative to R-only
conditioning, while R+S lowers RMSE for the surface-targeted group by
$13.31\%$.
When both sources are added, R+A+S lowers the all-variable,
surface-targeted, and upper-air temperature and wind metrics by $9.24\%$,
$14.17\%$, and $9.30\%$, respectively. Relative to the stronger of R+A and R+S within each
group, R+A+S provides additional reductions of $3.89$, $0.86$, and $2.56$
percentage points, respectively.
Figure~\ref{fig:frozen2019_composition_summary} compares these
three configurations at the variable-group level, while
Figure~\ref{fig:frozen2019_per_variable} reports the changes for each state
variable. Physical-unit annual means from the same
reconstructions are given in
Table~\ref{tab:absolute_rmse_full723}; for example, from R-only
to R+A+S, $t2m$ RMSE over the CONUS domain decreases from $1.725$ to $1.380$ K, while
$v500$ decreases from $3.042$ to $2.695\,\mathrm{m\,s^{-1}}$.

\begin{table}[t]
\centering
\footnotesize
\caption{Grouped mean RMSE changes
$\overline{\Delta}_{G,\mathcal K'}^{P}$ over 723 matched 2020 analysis times.
RMSE is evaluated against ERA5, and changes are relative to
R-only conditioning; negative values indicate lower RMSE.}
\label{tab:full_year_main}
\renewcommand{\arraystretch}{1.16}
\begin{tabular*}{1\textwidth}{@{\extracolsep{\fill}}lrrr@{}}
\toprule
Conditioning & All 13 & Surface-targeted &
\shortstack{Upper-air temperature\\and wind} \\
\midrule
\multicolumn{4}{@{}l}{\textit{CONUS domain}} \\
R+A   & $-4.46\%$ & $-2.28\%$  & $-6.74\%$ \\
R+S   & $-5.35\%$ & $-13.31\%$ & $-2.99\%$ \\
R+A+S & $\mathbf{-9.24\%}$ & $\mathbf{-14.17\%}$ & $\mathbf{-9.30\%}$ \\
\addlinespace[3pt]
\multicolumn{4}{@{}l}{\textit{Global}} \\
R+A   & $+0.006\%$ & $-0.018\%$ & $-0.002\%$ \\
R+S   & $-0.054\%$ & $-0.157\%$ & $-0.022\%$ \\
R+A+S & $-0.035\%$ & $-0.146\%$ & $-0.015\%$ \\
\addlinespace[3pt]
\multicolumn{4}{@{}l}{\textit{Global excluding CONUS domain}} \\
R+A   & $+0.030\%$ & $-0.003\%$ & $+0.034\%$ \\
R+S   & $-0.019\%$ & $-0.052\%$ & $-0.005\%$ \\
R+A+S & $+0.019\%$ & $-0.036\%$ & $+0.035\%$ \\
\bottomrule
\end{tabular*}
\end{table}

\begin{figure}[H]
    \centering
    \includegraphics[width=0.99\textwidth]
    {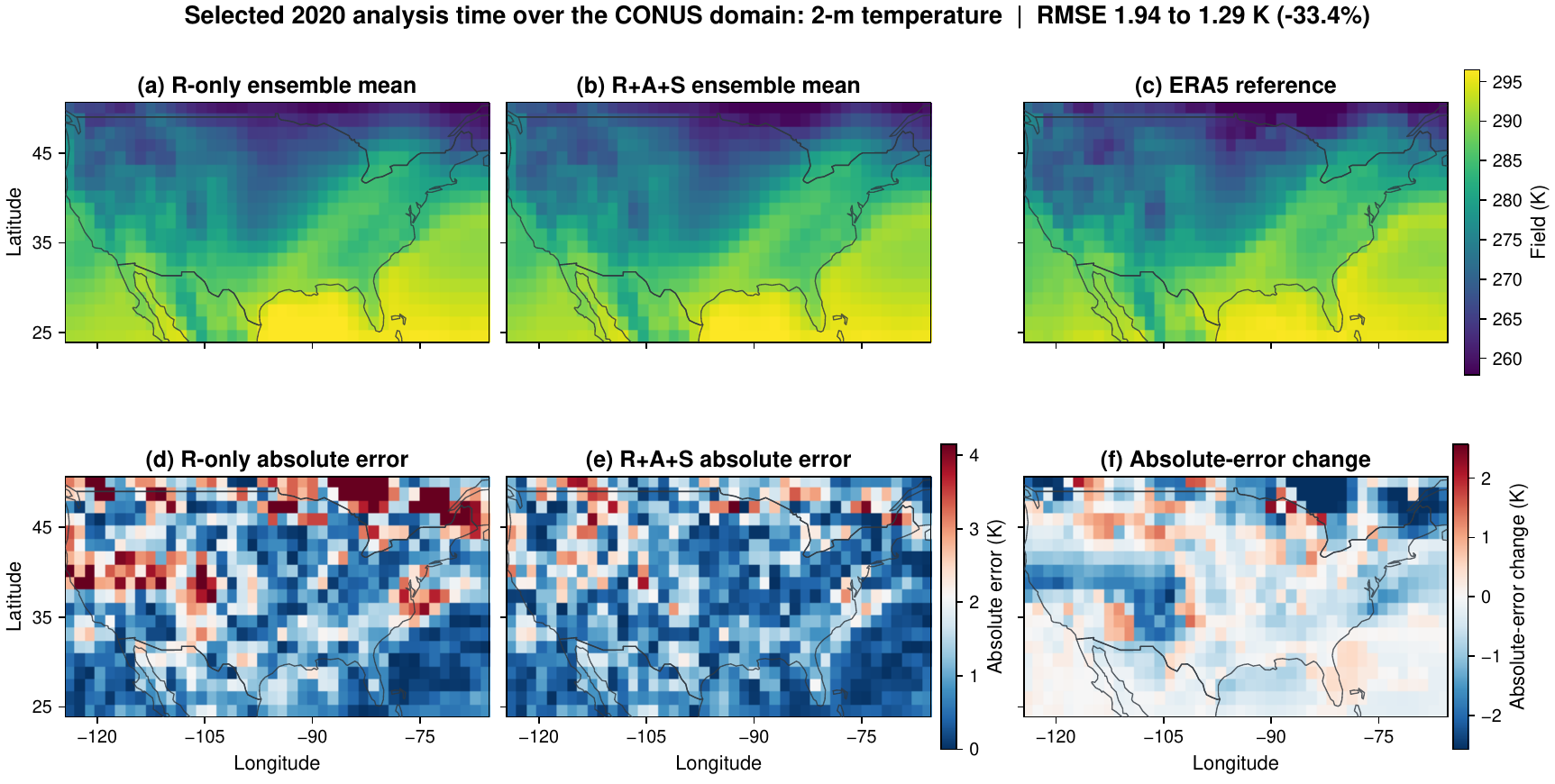}
    \caption{Selected 2020 analysis time for $t2m$ over the
    CONUS domain: (a) R-only ensemble mean, (b) R+A+S ensemble mean,
    (c) ERA5, (d) R-only absolute error, (e) R+A+S absolute error, and
    (f) their difference, R+A+S minus R-only. Panels (d) and (e) share a color scale capped at their pooled 97.5th
    percentile; blue and red indicate lower and higher absolute error,
    respectively. Panel (f) uses a symmetric scale capped at the 97.5th percentile
    of the absolute changes; blue indicates an error reduction and red an error
    increase. RMSE decreases from $1.94$ to $1.29$ K.}
    \label{fig:joint_high_skill_t2m}
\end{figure}

\begin{figure}[H]
    \centering
    \includegraphics[width=0.99\textwidth]
    {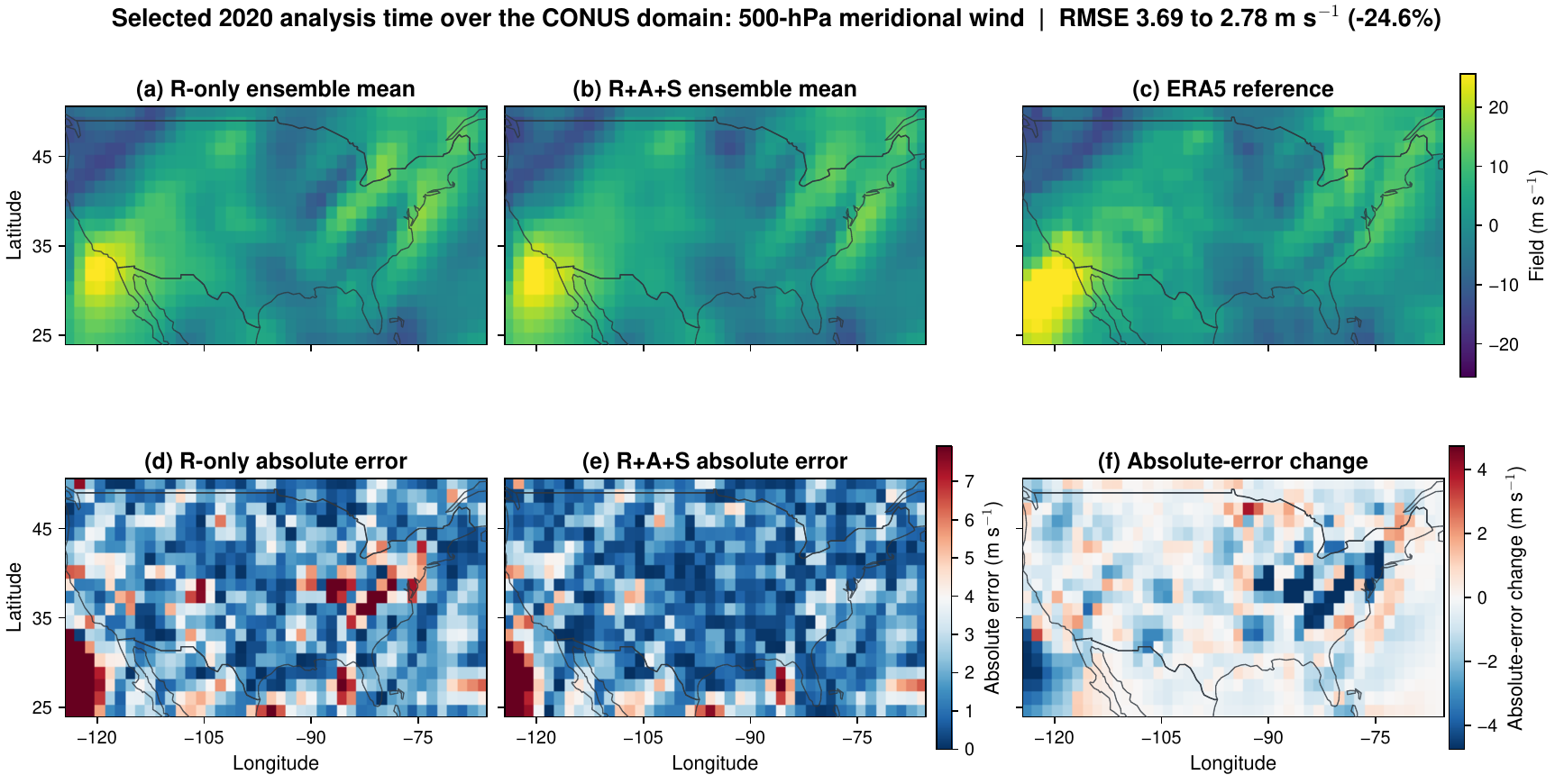}
    \caption{Selected 2020 analysis time for $v500$ over the
    CONUS domain for the same case and panel layout as
    Figure~\ref{fig:joint_high_skill_t2m}. RMSE decreases from $3.69$ to
    $2.78\,\mathrm{m\,s^{-1}}$.}
    \label{fig:joint_high_skill_v500}
\end{figure}

Figures~\ref{fig:joint_high_skill_t2m} and
\ref{fig:joint_high_skill_v500} show a selected 2020 analysis time for one
surface variable directly constrained by S and one upper-air variable directly
constrained by A. The analysis time was selected because R+A+S substantially
reduced RMSE relative to R-only conditioning for both displayed variables. For
this case, the R+A+S ensemble means are closer to ERA5, and the figures show
where their absolute errors differ from those of R-only conditioning. The
annual results above establish performance across all 723 analysis times.
Appendix Section~\ref{sec:supplemental_spatial_fields} further checks that the
lower errors are not accompanied by blanket smoothing over the CONUS domain.

\FloatBarrier

\subsection{Temporal Robustness under the 14-Day Moving-Block Bootstrap}

For R+A+S over the CONUS domain, the 14-day moving-block
bootstrap gives a mean RMSE change of $-9.24\%$ across all 13 variables, with
a 95\% interval of $[-9.57,-8.92]\%$. The corresponding intervals are
$[-14.76,-13.57]\%$ for the surface-targeted group and
$[-9.83,-8.80]\%$ for the upper-air temperature and wind group. All 13
variable-specific intervals over the CONUS domain are below zero;
Figure~\ref{fig:frozen2019_per_variable_bootstrap} and
Table~\ref{tab:per_variable_bootstrap_final} in the Appendix resolve this
result by state variable. Table~\ref{tab:block_bootstrap_final}
also records the global intervals and those for the global
domain outside CONUS; the mean change across all 13 variables
outside CONUS remains near zero.

\begin{table}[htbp]
\centering
\small
\caption{Grouped mean RMSE changes
$\overline{\Delta}_{G,\mathcal K'}^{\mathrm{R+A+S}}$ and 95\% intervals from
the 14-day moving-block bootstrap. Negative values indicate lower RMSE relative
to R-only conditioning.}
\label{tab:block_bootstrap_final}
\renewcommand{\arraystretch}{1.08}
\begin{tabular*}{\textwidth}{@{\extracolsep{\fill}}llrr@{}}
\toprule
Region & Variable group &
$\overline{\Delta}_{G,\mathcal K'}^{\mathrm{R+A+S}}$ (\%) &
95\% interval (\%) \\
\midrule
CONUS domain & All 13 variables & $\mathbf{-9.236}$ & $\mathbf{[-9.566,-8.918]}$ \\
             & Surface-targeted & $\mathbf{-14.173}$ & $\mathbf{[-14.759,-13.566]}$ \\
             & \shortstack[l]{Upper-air temperature\\and wind} & $\mathbf{-9.301}$ & $\mathbf{[-9.825,-8.799]}$ \\
\cmidrule(lr){1-4}
Global       & All 13 variables & $-0.0353$ & $[-0.0618,-0.0083]$ \\
             & Surface-targeted & $-0.1457$ & $[-0.1847,-0.1069]$ \\
             & \shortstack[l]{Upper-air temperature\\and wind} & $-0.0152$ & $[-0.0380,0.0080]$ \\
\cmidrule(lr){1-4}
Global excluding CONUS domain & All 13 variables & $+0.0188$ & $[-0.0078,0.0458]$ \\
              & Surface-targeted & $-0.0364$ & $[-0.0760,0.0025]$ \\
              & \shortstack[l]{Upper-air temperature\\and wind} & $+0.0352$ & $[0.0132,0.0579]$ \\
\bottomrule
\end{tabular*}
\end{table}

\subsection{Prediction at Held-Out Observations}
\label{sec:heldout_results}

\begin{figure}[tbp]
    \centering
    \includegraphics[width=1.0\textwidth]{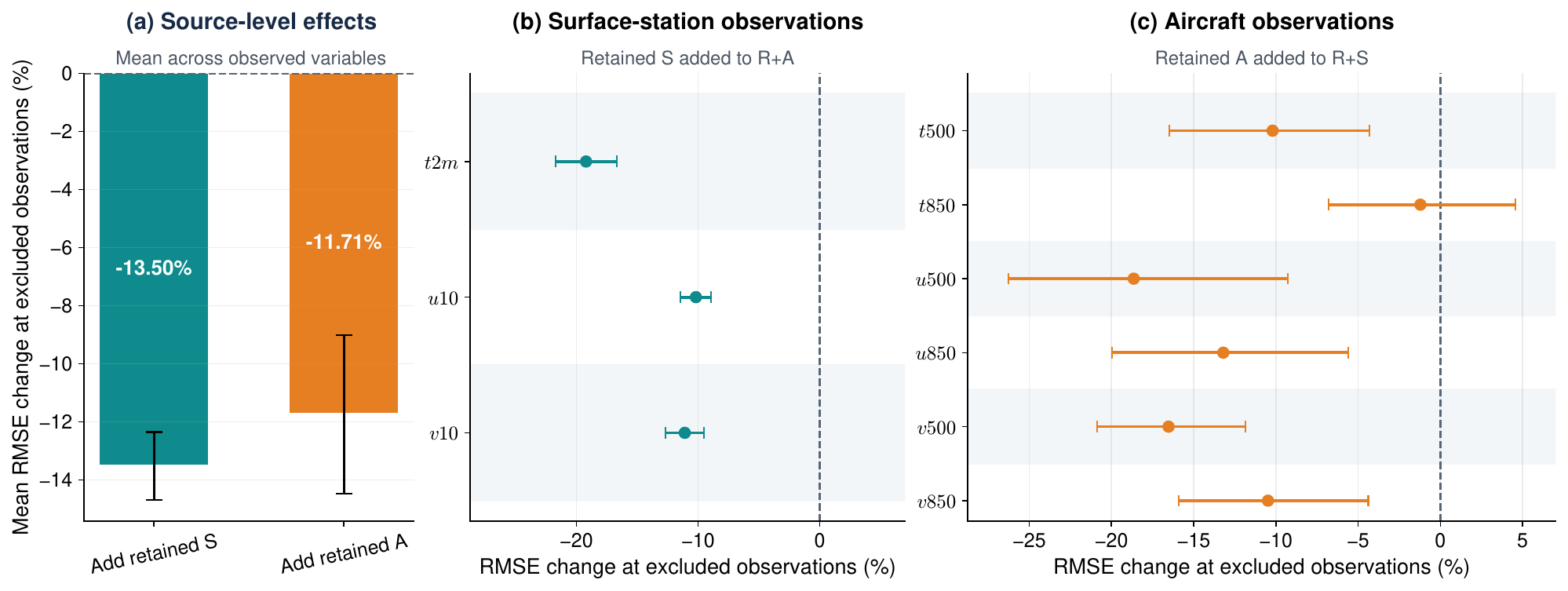}
    \caption{RMSE changes at CONUS observations excluded from
    conditioning across 24 seasonally distributed 2020 analysis times. Panel (a)
    reports source-level means when the retained 80\% of S is added to R+A and the
    retained 80\% of A is added to R+S; panels (b) and (c) report
    variable-specific changes. Error bars show 95\% paired-analysis-time bootstrap
    intervals, and negative values indicate lower RMSE.}      
    \label{fig:holdout_matched_marginal}
\end{figure}

Table~\ref{tab:holdout_matched_marginal} compares prediction
errors at CONUS observations excluded from conditioning. Adding the retained 80\%
of S-observed cells to R+A lowers RMSE at the excluded S observations by
$13.50\%$, with a 95\% paired-analysis-time bootstrap interval of
$[-14.68,-12.35]\%$ and lower error at all 24 analysis times. Adding the
retained 80\% of A cell--variable targets to R+S lowers RMSE at the excluded A
observations by $11.71\%$, with an interval of $[-14.48,-9.00]\%$ and lower
error at 23 of 24 analysis times. All three S-variable intervals lie below
zero. Five of the six A-variable intervals lie below zero; the $t850$ mean is
$-1.21\%$, but its interval crosses zero.
Figure~\ref{fig:holdout_matched_marginal} presents the same
comparisons as source-level means and separately for each observed variable.

These results show that S improves prediction at excluded S
observations beyond R+A, while A improves prediction at excluded A
observations beyond R+S. Because the excluded observations come from the same
surface-station and aircraft observing systems as the retained observations,
this evaluation measures interpolation within those systems rather than
transfer to a new observing system.

\begin{table}[tbp]
\centering
\caption{RMSE changes at 2020 CONUS observations excluded from conditioning
across 24 seasonally distributed analysis times. For each source, 80\% of the targets
are retained for conditioning and the remaining 20\% are used for evaluation.
Intervals are 95\% paired-analysis-time bootstrap intervals.}
\label{tab:holdout_matched_marginal}
\small
\setlength{\tabcolsep}{5pt}
\renewcommand{\arraystretch}{1.18}
\begin{tabularx}{\textwidth}{@{}l>{\raggedright\arraybackslash}Xrrr@{}}
\toprule
Evaluated source & Conditioning comparison & Mean change & 95\% interval &
Lower-RMSE cases \\
\midrule
Surface station (S) & Retained 80\% of S added to R+A &
$-13.50\%$ & $[-14.68,-12.35]\%$ & $24/24$ \\
Aircraft (A) & Retained 80\% of A added to R+S &
$-11.71\%$ & $[-14.48,-9.00]\%$ & $23/24$ \\
\bottomrule
\end{tabularx}
\end{table}

\subsection{Probabilistic Evaluation}
\label{sec:probabilistic_results}

Table~\ref{tab:frozen2019_probabilistic} evaluates whether the
ensemble-mean RMSE reductions extend to the ensemble distribution using empirical CRPS.
\begin{table}[tbp]
\centering
\caption{Grouped mean percentage changes in ensemble-mean
RMSE and empirical CRPS over the CONUS domain for R+A+S relative to R-only
conditioning, evaluated with 16-member ensembles at 723 matched 2020 analysis
times. Both scores are evaluated against ERA5. Brackets give 95\% intervals
for the CRPS changes from the 14-day moving-block bootstrap.}
\label{tab:frozen2019_probabilistic}
\small
\renewcommand{\arraystretch}{1.22}
\begin{tabular*}{\textwidth}{@{\extracolsep{\fill}}lcc@{}}
\toprule
Variable group & RMSE change & CRPS change [95\% interval] \\
\midrule
All 13 variables
& $-9.24\%$ & $-9.98\%\;[-10.40,-9.57]$ \\
Surface-targeted variables
& $-14.17\%$ & $-15.01\%\;[-15.75,-14.27]$ \\
Upper-air temperature and wind variables
& $-9.30\%$ & $-10.45\%\;[-11.11,-9.83]$ \\
\bottomrule
\end{tabular*}
\end{table}

Over the CONUS domain, R+A+S reduces CRPS relative to R-only
conditioning by $9.98\%$ across all
13 variables, $15.01\%$ for the surface-targeted group, and $10.45\%$ for the
upper-air temperature and wind group. 
All three 95\% intervals from the 14-day moving-block bootstrap remain below zero (Table~\ref{tab:frozen2019_probabilistic}). Using the finite-ensemble fair-CRPS estimator described in
Appendix~\ref{sec:metric_definitions} changes these means by at most
$0.10$ percentage points and leaves the ranking unchanged. In
the global domain outside CONUS, the mean CRPS change across all 13 variables
is $+0.004\%$
with a 95\% interval of
$[-0.024,+0.033]\%$, consistent with the near-neutral deterministic result.

Consistent reductions in empirical CRPS, fair CRPS, and ensemble-mean RMSE show that the annual improvement extends beyond ensemble means to
probabilistic skill. Supplemental finite-ensemble dispersion diagnostics are
reported in Appendix~\ref{sec:supplemental_validation}.

\section{Mechanism Diagnostics: From Source Likelihood
Gradients to Runtime Guidance}
\label{sec:mechanism_diagnostics}

\begin{figure}[H]
    \centering
    \includegraphics[width=0.99\textwidth]
    {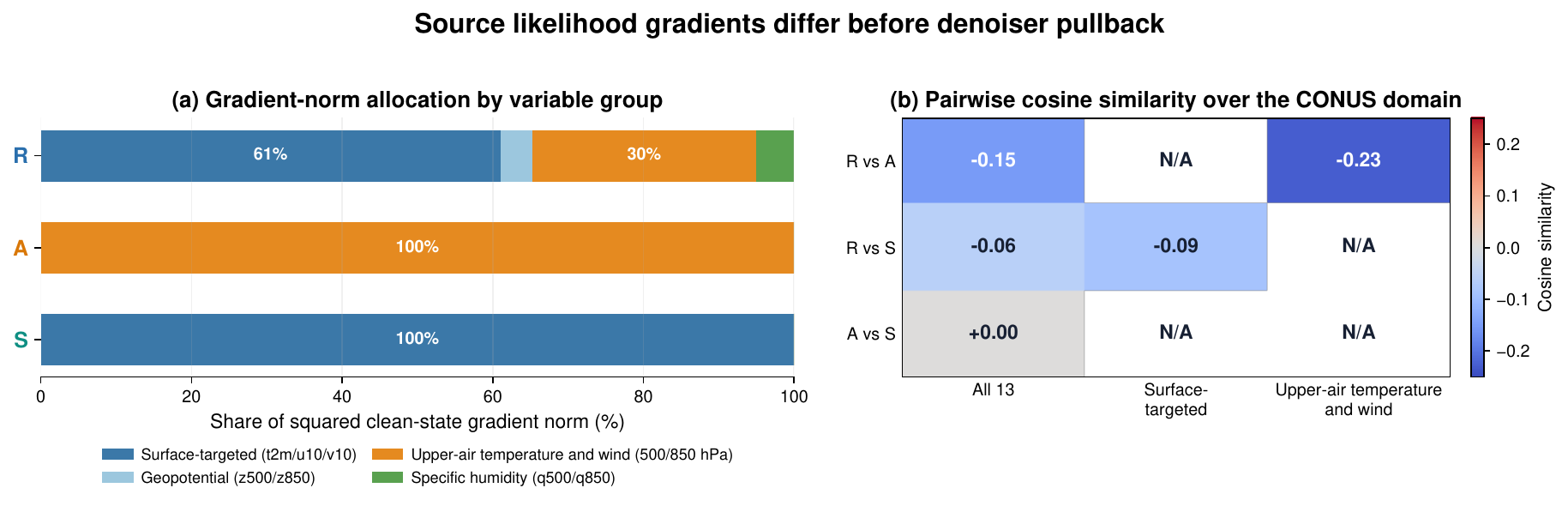}
    \caption{Source-specific clean-state likelihood-gradient geometry for one
    2020 case before denoiser-Jacobian pullback. Panel (a) shows the fraction
    of each source's full-grid squared gradient norm in each variable group.
    Panel (b) shows pairwise cosine similarity over the CONUS domain for all
    13 variables and for the
    surface-targeted group and upper-air temperature and wind group.
    N/A indicates that at least one projected gradient has zero norm over the
    selected variable subset, so the cosine similarity is undefined.}
    \label{fig:frozen2019_static_mechanism}
\end{figure}

\begin{figure}[tbp]
    \centering
    \includegraphics[width=0.99\textwidth]
    {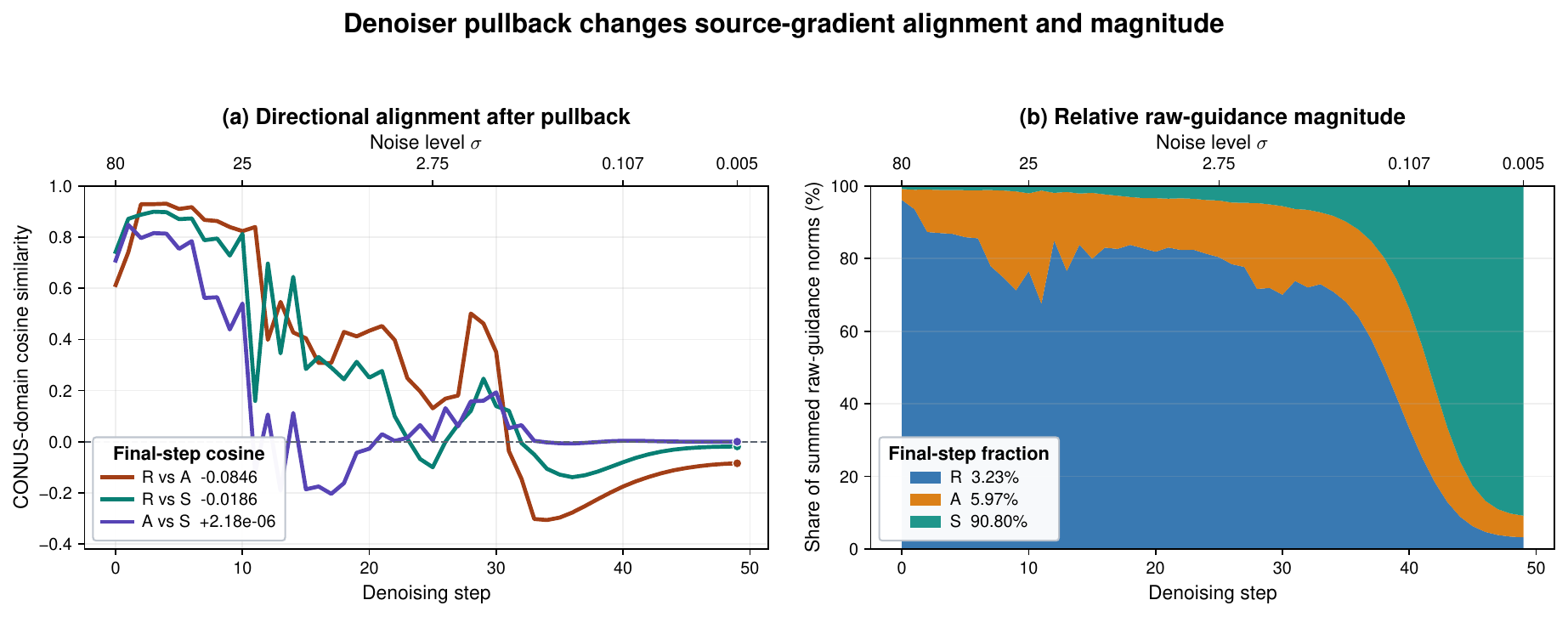}
    \caption{Runtime interaction of the source-specific likelihood gradients
    along one reverse trajectory. Panel (a) shows their pairwise cosine
    similarities over the CONUS domain after denoiser-Jacobian pullback.
    Panel (b) shows each source's raw-guidance norm as a fraction of the sum
    of the three norms over all 13 variables on the global model grid. Both
    diagnostics are computed before the shared summation and clipping
    operation.}
    \label{fig:frozen2019_runtime_mechanism}
\end{figure}

\begin{figure}[tbp]
    \centering
    \includegraphics[width=0.99\textwidth]
    {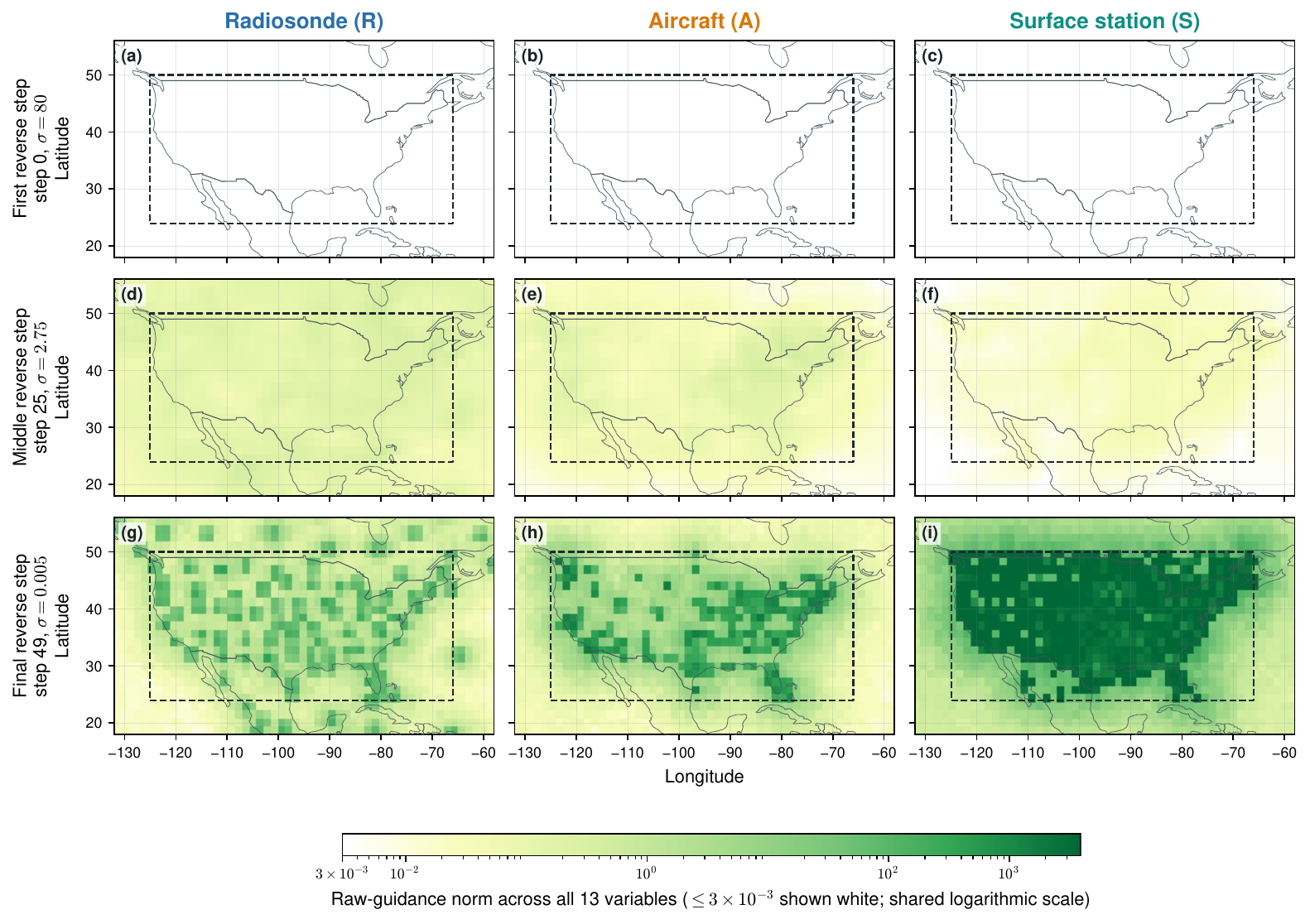}
    \caption{Spatial norms of the source-specific raw runtime guidance along
    the selected reverse trajectory. Columns show radiosonde (R), aircraft
    (A), and surface-station (S) contributions, and rows show the first,
    middle, and final reverse steps. All panels use the same logarithmic color scale, 
    with values at or below $3\times10^{-3}$ shown in white and the upper limit set by the
    pooled 99.9th percentile. Dashed rectangles mark the CONUS domain. 
    The norms are computed before the shared summation and clipping operation.}
    \label{fig:frozen2019_runtime_footprints}
\end{figure}

The annual results show that adding A and S to R-only
conditioning produces upper-air and surface improvements that are both
retained in R+A+S. We therefore examine how the source-specific likelihood
gradients interact within the sampler. Specifically, we ask whether the A and
S gradients act on distinct variables before denoiser pullback, how the shared
denoiser changes their alignment and relative magnitudes during sampling, and
where the resulting runtime guidance is distributed spatially.
Figures~\ref{fig:frozen2019_static_mechanism},
\ref{fig:frozen2019_runtime_mechanism}, and
\ref{fig:frozen2019_runtime_footprints} evaluate these questions for one 2020
case. These figures provide mechanism diagnostics for that case, while the
annual and held-out evaluations reported above provide the evidence for
performance across analysis times.

We first evaluate the clean-state likelihood gradients at the
ensemble-mean R+A+S reconstruction. Let
$\mathbf{x}_{0,\star}^{(e)}$ denote member $e$ of the 16-member R+A+S ensemble
for this case and define
\begin{equation}
    \overline{\mathbf{x}}_{0,\star}
    =\frac{1}{\Nens}\sum_{e=1}^{\Nens}\mathbf{x}_{0,\star}^{(e)},
    \qquad
    \mathbf d_m^{\star}
    =\mathbf d_m(\overline{\mathbf{x}}_{0,\star};\sigma_\star),
    \qquad \sigma_\star=0.005.
    \label{eq:static_direct_guidance_case}
\end{equation}
Here, $\sigma_\star=0.005$ is the lowest nonzero noise level
in the EDM sampling schedule used throughout the study, and $\mathbf d_m$ is
the clean-state likelihood gradient defined in
Eq.~\eqref{eq:direct_modality_guidance}. 
For a spatial domain $G$ and variable
subset $\mathcal K'\subseteq\mathcal K$, let $\Pi_{G,\mathcal K'}$ select the
corresponding entries of this gradient. We define its squared magnitude and
the cosine similarity between sources $m$ and $n$ as
\begin{equation}
\begin{aligned}
    Q_m(G,\mathcal K')
    &=\left\|\Pi_{G,\mathcal K'}\mathbf d_m^{\star}\right\|_2^2,\\
    \rho_{mn}(G,\mathcal K')
    &=\frac{\left\langle \Pi_{G,\mathcal K'}\mathbf d_m^{\star},
    \Pi_{G,\mathcal K'}\mathbf d_n^{\star}\right\rangle}
    {\left\|\Pi_{G,\mathcal K'}\mathbf d_m^{\star}\right\|_2
     \left\|\Pi_{G,\mathcal K'}\mathbf d_n^{\star}\right\|_2}.
\end{aligned}
\label{eq:static_direct_guidance_geometry}
\end{equation}
$Q_m(G,\mathcal K')$ measures the squared gradient magnitude over the
selected domain and variables, while $\rho_{mn}(G,\mathcal K')$ measures
directional alignment. We report $\rho_{mn}(G,\mathcal K')$ as N/A whenever
either projected gradient has zero norm over the selected domain and variable
subset, because the cosine similarity is then undefined. Panel (a) reports the fraction of each source's
full-grid $Q_m$ assigned to each variable group. Because these fractions are
normalized within each source and cosine similarity is invariant to positive
scaling, the positive scalar $\lambda_m/v_m(\sigma_\star)$ does not affect the
displayed values. No denoiser-Jacobian pullback has yet been applied.
We partition the 13 state variables into four groups:
surface-targeted $(t2m,u10,v10)$, geopotential $(z500,z850)$,
upper-air temperature and wind $(t500,u500,v500,t850,u850,v850)$,
and specific humidity $(q500,q850)$. In Figure~\ref{fig:frozen2019_static_mechanism}(a), A assigns
100\% of its squared clean-state gradient norm to the six upper-air
temperature and wind variables, while S assigns 100\% to the three
surface-targeted variables. Panel (b)
evaluates $\rho_{mn}$ over the CONUS domain. The all-variable A--S cosine is
zero because A and S act directly on disjoint variables. Across the displayed
variable groups, the nonzero cosines involving R have magnitudes no greater
than $0.23$ in this case.

\FloatBarrier

We next follow one ensemble member along its reverse
trajectory. After denoiser-Jacobian pullback, the raw runtime guidance
$\mathbf g_{m,i}^{\mathrm{raw}}=J_{D,i}^{\top}\mathbf d_m$ is expressed in
the noisy sampler state. Figure~\ref{fig:frozen2019_runtime_mechanism}(a)
evaluates the same pairwise cosine similarity using
$\mathbf g_{m,i}^{\mathrm{raw}}$ instead of $\mathbf d_m^\star$. Panel (b)
reports each source's fraction of the summed raw-guidance norms,
\begin{equation}
    \eta_{m,i}=\frac{\lVert\mathbf g_{m,i}^{\mathrm{raw}}\rVert_2}
    {\sum_{n\in\Mset}\lVert\mathbf g_{n,i}^{\mathrm{raw}}\rVert_2}.
    \label{eq:runtime_mechanism_summaries}
\end{equation}
Here, each norm is evaluated over all 13 variables on the global model grid.
For this trajectory, the largest pairwise cosine reaches
$0.93$ at high noise, showing that the shared denoiser can map gradients with
distinct direct variable support into closely aligned noisy-state directions.
The three pairwise cosines move back toward zero near the final step, when S
accounts for $90.80\%$ of the sum of the three raw-guidance norms. Thus, both
the directional alignment and relative magnitude of the source-specific
guidance change during sampling.

To complement these trajectory-level summaries,
Figure~\ref{fig:frozen2019_runtime_footprints} shows where the source-specific
raw runtime guidance is distributed at the first, middle, and final reverse
steps. The shared logarithmic scale preserves magnitude comparisons across
observation sources and reverse steps.

Together, these diagnostics show how source likelihood gradients that act
directly on different state variables can become coupled through the shared
denoiser in the selected case. The cosine similarities and norm fractions
quantify their directional and magnitude relationships during sampling, while
Figure~\ref{fig:frozen2019_runtime_footprints} shows the corresponding spatial
distributions. On the shared scale, the raw-guidance magnitudes are much
smaller at the first reverse step than near the end of the trajectory.
Because all three diagnostics are computed before the shared summation and
clipping operation, they do not decompose the final applied correction among
the three sources. They describe one selected trajectory and should not be
interpreted as annual mechanism statistics.

\FloatBarrier

\section{Discussion}

The central result is that one pretrained atmospheric diffusion prior can
incorporate radiosonde, aircraft, and surface-station observations through
source-specific interfaces without retraining the prior. These interfaces are
not interchangeable. The retained observations, state mapping,
residual-counting rule, and likelihood weighting determine how each source
constrains the common gridded state. 
Because the interface choices were fixed before the 2020
evaluation, the reported improvements do not reflect tuning to the evaluation
year.

Framed as generative atmospheric super-resolution, the results
show that the connection between heterogeneous in situ observations and a
gridded atmospheric prior is part of the inference problem.
Spatial resolution alone does not describe this connection
because in situ observation sources also differ in measured variables and
sampling density.
The observation interface makes these
differences explicit while allowing the same pretrained prior to be reused.

\subsection{Source-Specific Improvements and Shared-State Coupling}

The annual effects of A and S follow the variables that they observe directly.
R+A produces its larger reduction for the upper-air temperature and wind group, while R+S
produces its larger reduction for the surface-targeted group. R+A+S retains
both patterns and reduces RMSE for all 13 state variables over the CONUS domain.
Relative to the stronger of R+A and R+S in each group, R+A+S provides further
reductions of $3.89$, $0.86$, and $2.56$ percentage points for the
all-variable group, surface-targeted group, and
upper-air temperature and wind group, respectively.
The $9.24\%$ all-variable reduction therefore reflects information from both A
and S rather than the effect of either source alone.

The annual per-variable results further show that the response is structured
rather than uniform. The largest reduction occurs for $t2m$; wind and
geopotential variables also improve, while humidity changes are smaller.
Figures~\ref{fig:joint_high_skill_t2m} and
\ref{fig:joint_high_skill_v500} illustrate how lower errors appear spatially
for one selected analysis time, but the annual results provide the evidence
that these reductions persist across 2020. The single-case diagnostics in
Section~\ref{sec:mechanism_diagnostics} provide a computational explanation:
A and S act directly on disjoint variables, while the shared denoiser maps
their clean-state likelihood gradients into coupled runtime directions. This
diagnostic demonstrates a coupling pathway within the pretrained multivariate
prior, but does not imply that the same gradient alignment occurs at every
analysis time.

\subsection{Why Explicit Observation Interfaces Matter}

The 2019 aircraft-source comparison shows that adding more reports is not
necessarily beneficial. Reporting streams that supplied no or only sparse
eligible observations did not materially change reconstruction errors, while
including TAMDAR increased them. The observations retained from a source are
therefore part of the inference design rather than a detail that can always be
absorbed into a generic error scale.

The comparison of residual representations gives the same lesson for sampling
density. Averaging A and S reports within model-grid cells and assigning one
residual per occupied cell and variable performed better than counting every
report separately under the tested likelihood settings. Additional reports
can refine the cell-level target, but their number should not by itself give a
dense flight route, airport, or station cluster greater likelihood influence.

The methodological contribution is the separation between a reusable
atmospheric prior and explicit source-specific inference design. It does not
lie in any one observation operator or averaging rule in isolation, but in
their organization into interfaces that state which observations are retained,
how they are mapped to the gridded state, how their residuals are counted, and
how their likelihood contributions are weighted.
These choices can be inspected and modified separately for each source. A new
observation source can therefore be incorporated by defining these components,
calibrating them without changing the pretrained prior, and evaluating their
interaction with the existing interfaces.

\subsection{What the Evaluation Establishes}

The evaluation provides several distinct checks on the annual result. The intervals from the 14-day moving-block bootstrap remain below zero for
all three grouped RMSE changes and all 13 per-variable RMSE changes over the CONUS domain, showing
that the reductions are not carried by a few isolated analysis times. The $9.98\%$ all-variable CRPS reduction relative to R-only conditioning shows that the improvement extends beyond ensemble-mean RMSE to
marginal probabilistic skill.

The held-out evaluation addresses a different question. Adding the
retained S observations to R+A reduces RMSE by $13.50\%$ at excluded CONUS S
cells, while adding the retained A observations to R+S reduces RMSE by
$11.71\%$ at excluded CONUS A cell--variable targets. These results show interpolation to
observations excluded from conditioning within the same aircraft and
surface-station systems. They do not evaluate transfer to a new observing
system.

The geographic pattern also follows the direct support of A and S. The all-variable RMSE reduction relative to R-only conditioning is $9.24\%$ over the CONUS domain, whereas the
aggregate change in the global domain outside CONUS is near zero and its
14-day interval includes zero. The evidence therefore supports improved
reconstruction where A and S directly constrain the state, not a claim of a
global improvement from these regional interfaces.

\subsection{Limitations and Future Directions}

The modeling conclusions are conditioned on one pretrained 13-variable ERA5
prior and one set of sampler settings. Aircraft observations are assigned only
to the available 500 and 850 hPa state levels using fixed pressure windows.
The likelihood uses weighted squared residuals and does not explicitly model
correlated errors across observation sources or among nearby observations
beyond within-cell averaging. The shared post-sum clipping operation also
makes the applied correction nonlinear when the combined guidance saturates.
Finally, the mechanism diagnostics examine one analysis time and one reverse
trajectory rather than the full annual evaluation.

The evaluation has corresponding limits. 
ERA5 serves as the reference state for the interface
comparisons using the 2019 development cases and for the independent 2020
gridded evaluation; the corresponding deterministic, probabilistic, and
spatial results are therefore defined relative to the reanalysis.
A and S directly constrain only the
CONUS domain, and the held-out evaluation uses one random 80/20 split over 24
analysis times. The conditioning comparisons isolate the contributions of A
and S within the pretrained prior and sampler used here, but do not establish
superiority over classical data-assimilation systems or other diffusion-based
approaches.
Future work should test additional vertical levels and observation sources,
likelihoods with correlated errors, repeated or spatially structured holdouts,
globally distributed interfaces, alternative atmospheric priors and samplers,
and controlled comparisons with established reconstruction methods.

\section{Conclusion}

We formulate the reconstruction of a regularly gridded 13-variable
atmospheric state from radiosonde, aircraft, and surface-station observations
as generative atmospheric super-resolution. 
A pretrained diffusion model supplies the atmospheric prior.
Each observation source is connected to this prior through an
interface that defines which measurements are retained and how they are mapped
to the gridded state.
The interfaces also determine how residuals are aggregated and
weighted during inference.
This design allows observations with different sampling
geometries and densities to be combined without retraining the prior.

We selected the A and S interface designs and calibrated their
likelihood parameters using the 24 prespecified 2019 development cases, then
evaluated the selected interfaces at 723 matched analysis times in 2020
without further tuning.
Relative to R-only conditioning, R+A+S lowers ERA5-referenced
RMSE across all 13 state variables over the CONUS domain, with a mean
reduction of $9.24\%$.
Across the same 2020 evaluation cases, R+A+S also lowers
ERA5-referenced CRPS.
Separate held-out evaluations show reduced errors at aircraft
and surface-station targets excluded from conditioning.

These results demonstrate
that explicit, composable observation interfaces provide a practical route
for reusing a pretrained atmospheric generative prior across heterogeneous in situ observation sources while keeping the connection between each observation source
and the gridded state inspectable.

\section*{Acknowledgments}

Computations were performed using resources provided by Purdue University's
Rosen Center for Advanced Computing. Training of the pretrained atmospheric
diffusion model used in this work was performed on Perlmutter using resources
of the National Energy Research Scientific Computing Center (NERSC), a U.S.
Department of Energy Office of Science User Facility. The authors acknowledge
NOAA for the
Integrated Global Radiosonde Archive and Meteorological Assimilation Data
Ingest System observations, and the Copernicus Climate Change Service for the
ERA5 reanalysis. This work was supported in part by the Defense Advanced
Research Projects Agency (DARPA) under Award No.~HR0011-26-3-E050 (program point of contact:
Dr. Yannis Kevrekidis) and ARO Young Investigator Award W911NF-24-1-0315 (program point of contact: Dr. Robert Martin).
The funders played no role in study design, data collection, analysis and
interpretation of data, or the writing of this manuscript.

\section*{Author Contributions}

\noindent\textbf{Y.X.:} Conceptualization, methodology, software, validation,
formal analysis, investigation, data curation, visualization, and
writing--original draft.

\noindent\textbf{D.C.:} Conceptualization and guidance
on experimental design and observational dataset selection.

\noindent\textbf{H.G.:} Methodology and software, including contributions to
the development and implementation of the original computational framework.

\noindent\textbf{S.W.:} Conceptualization and guidance on observational dataset
selection, manuscript framing, discussion, and interpretation of results.

\noindent\textbf{R.M.:} Conceptualization, supervision, project administration,
funding acquisition, and writing--review and editing.

All authors reviewed and approved the final manuscript.

\section*{AI Use Disclosure}

The language of the manuscript was partially improved using large language
models; all content generated by these tools was carefully reviewed and
verified by the authors before inclusion.

\section*{Competing Interests}

The authors declare no competing interests.

\section*{Data Availability}

IGRA observations are publicly available from NOAA's National Centers for
Environmental Information \citep{noaa_ncei_igra}. MADIS aircraft and
surface-station observations are distributed by NOAA
\citep{noaa_madis_aircraft,noaa_madis_surface}, and ERA5 is available through
the Copernicus Climate Data Store \citep{hersbach2020era5}.

The processed data products and pretrained model supporting the findings of
this study are available upon reasonable request from the authors.

\section*{Code Availability}

Research code and compact numerical summaries are publicly available in the
\href{https://github.com/ISCLPurdue/generative_multimodal_atmospheric_super_resolution}{project GitHub repository}.
The repository also contains the selected interface settings, manifests for
the 2020 evaluation cases, and the processing, sampling, and evaluation code
used in this study.

\FloatBarrier

\bibliographystyle{unsrtnat}
\bibliography{references}

\FloatBarrier
\appendix
\clearpage

\section{Sampling Parameters and Selected Analysis Times}
\label{sec:run_lineage}

The matched member seeds used across conditioning configurations are derived
from a base seed of 17. The EDM schedule uses $\sigma_{\min}=0.005$,
$\sigma_{\max}=80$, and $\rho=7$.

The equal-cell aircraft example in
Figure~\ref{fig:equal_cell_operator} and the clean-state and runtime mechanism
diagnostics in Figures~\ref{fig:frozen2019_static_mechanism},
\ref{fig:frozen2019_runtime_mechanism}, and
\ref{fig:frozen2019_runtime_footprints} use 2020-01-01 00 UTC. The field-level reconstruction examples
in Figures~\ref{fig:joint_high_skill_t2m} and
\ref{fig:joint_high_skill_v500} use 2020-03-10 12 UTC. These dates identify
the selected single-case examples; the quantitative conclusions are based on
the annual and held-out evaluations described in the main text.

\FloatBarrier

\section{Observation Processing and Input Stability}
\label{sec:observation_audit}

\subsection{Aircraft Data Coverage, Timing, and Quality Control}

The processed 2020 MADIS aircraft archive contains one aircraft-observation
file for each of the 732 nominal 00 and 12 UTC analysis times in the leap year.
This count describes aircraft-data availability; the annual evaluation uses
the 723 analysis times for which all required inputs are available, as defined
in Section~\ref{sec:independent_year_development}. Across the 732 files, the archive
contains 15,850,004 raw reports. Every reported observation time is finite and
falls within the hour beginning at the nominal file timestamp. Aircraft
reports are associated with the ERA5 state at that nominal hour without
sub-hour interpolation.

All files contain the MADIS quality-control fields required to screen
observation time, position, altitude, temperature, and wind. Among the 889,024
reports remaining after reporting-system selection and pressure-level
matching, every report passes the MADIS observation-time check. Applying that check as a filter would therefore remove no additional reports. No report used by the final A interface relies on a fallback for
missing quality-control information.

\subsection{Monthly Differences between Observations and ERA5}

\begin{figure}[!htbp]
    \centering
    \includegraphics[width=0.8\textwidth]
    {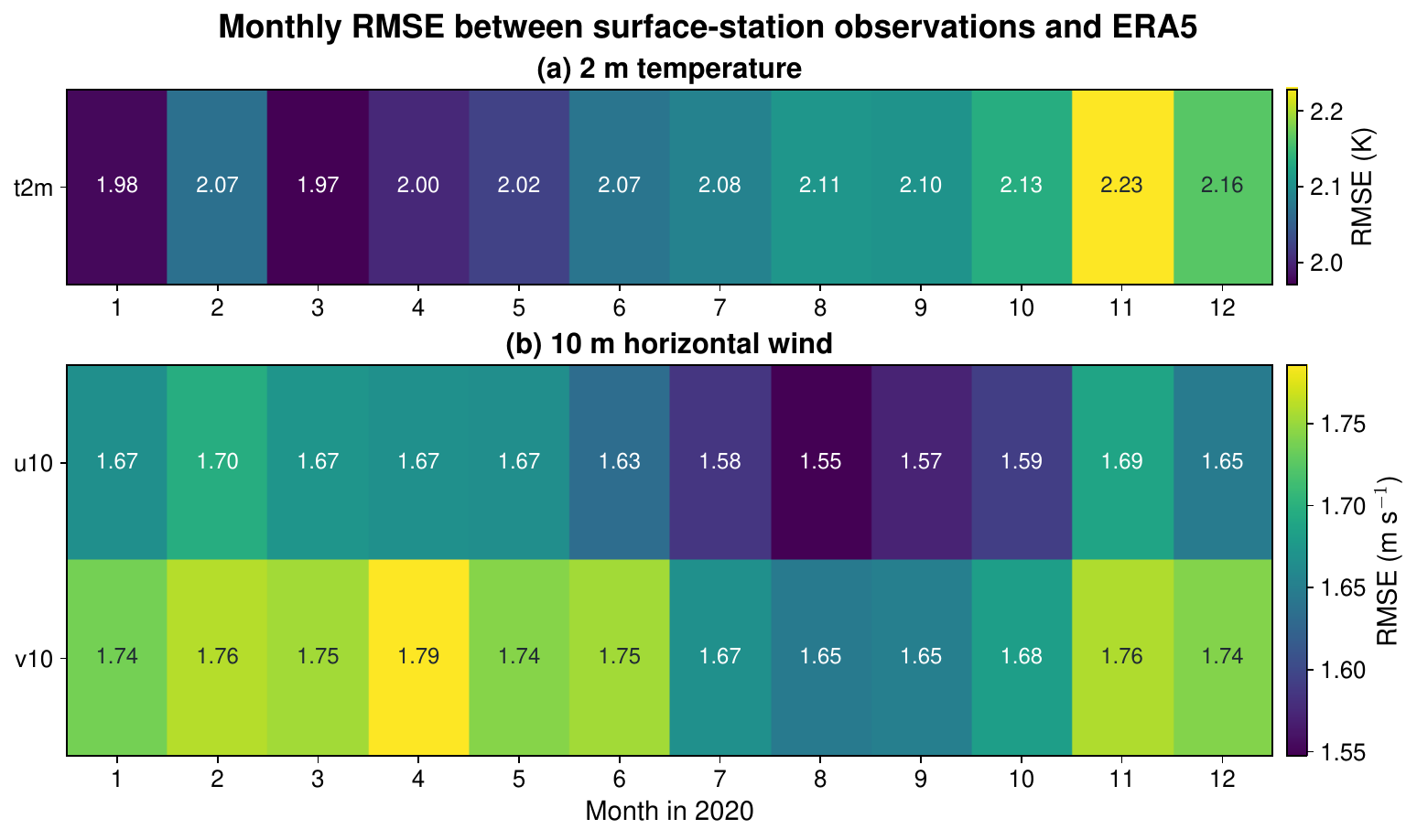}
    \caption{Monthly mean of the per-analysis-time RMSE between retained MADIS
    METAR surface-station observations and the corresponding ERA5 values over
    the CONUS domain: (a) $t2m$ and (b) $u10$ and $v10$. The observations
    satisfy the $\pm60$-minute temporal-matching, quality-control, and
    physical-range criteria.}
    \label{fig:metar_monthly_obs_error}
\end{figure}

\begin{figure}[!htbp]
    \centering
    \includegraphics[width=0.8\textwidth]
    {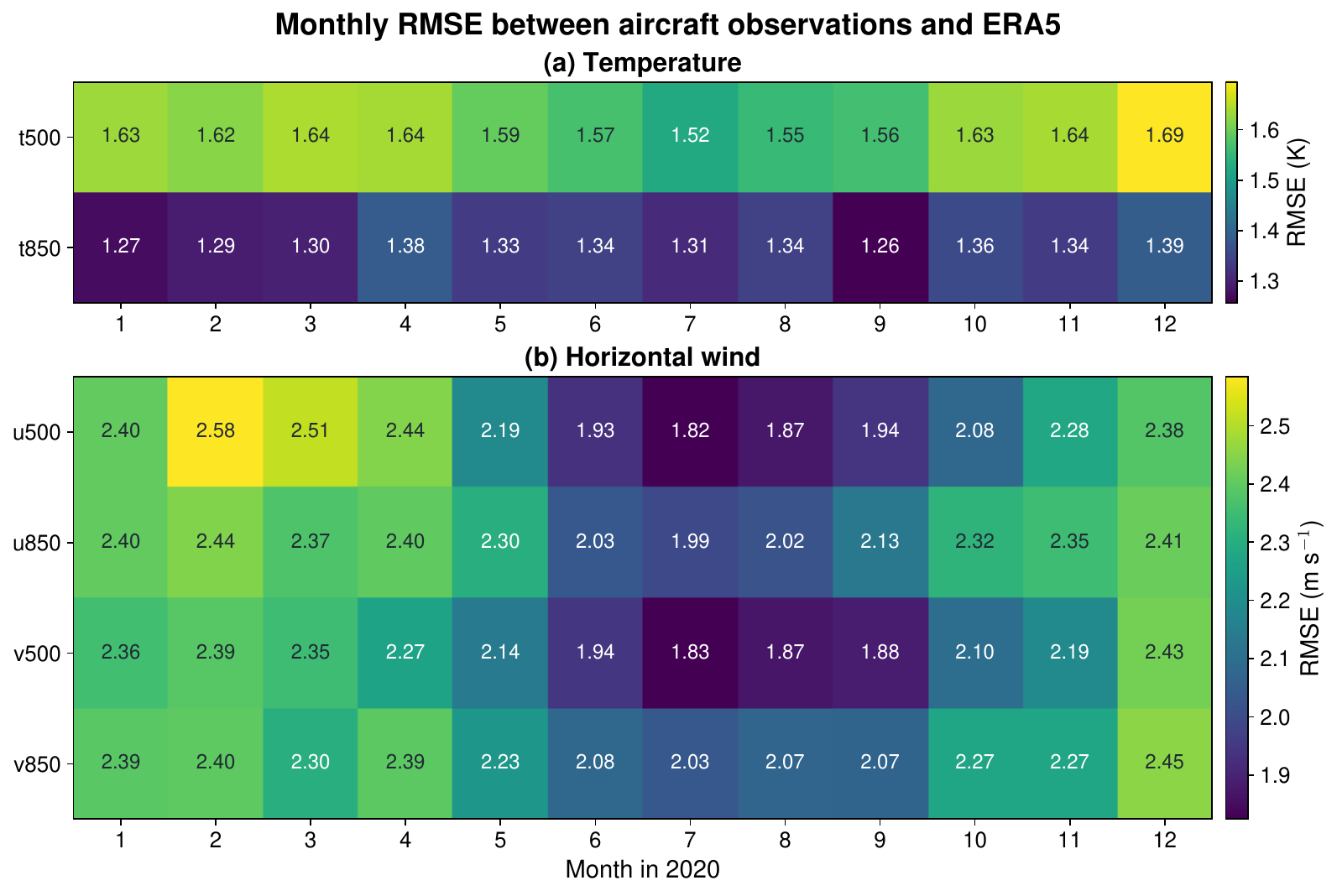}
    \caption{Monthly mean of the per-analysis-time RMSE between retained MADIS
    ABO aircraft observations and the corresponding ERA5 values over the
    CONUS domain: (a) $t500$ and $t850$ and (b) $u500$, $u850$, $v500$, and
    $v850$. The observations satisfy the final reporting-system selection,
    quality-control criteria, and $\pm25$-hPa pressure-level matching.}
    \label{fig:abo_monthly_obs_error}
\end{figure}

Figures~\ref{fig:metar_monthly_obs_error} and
\ref{fig:abo_monthly_obs_error} report monthly mean RMSE between the retained
observations and the corresponding ERA5 values after the final observation
processing. The monthly values vary but remain within comparable overall
ranges throughout 2020, with no evident month-specific failure in the retained
inputs. These figures assess the stability of the processed observation
sources; they are not posterior-performance metrics and were not used to
retune the interfaces during the
independent 2020 evaluation.

\FloatBarrier

\section{Interface Selection and Calibration Details}
\label{sec:calibration_inventory}

This section provides the 2019 comparisons underlying the interface designs
and likelihood parameters summarized in Table~\ref{tab:final_protocols}.
Every candidate is evaluated on the same 24 prespecified
2019 development cases.
All gridded RMSE values used during development are computed
against ERA5, and RMSE changes are reported relative to R-only conditioning
unless stated otherwise.
The separate A and S comparisons consider both the variables directly
constrained by that source and all 13 state variables. The final R+A+S
selection uses the all-variable RMSE change as its primary score and checks
the surface-targeted group and upper-air temperature and wind group as secondary outcomes.
Table~\ref{tab:calibration_inventory} summarizes the three
development stages and their selected results.

\begin{table}[htbp]
\centering
\small
\caption{Summary of the three stages used to select the A and
S interface designs and likelihood parameters from the 24 prespecified 2019
development cases. All reported RMSE changes are computed against ERA5 and
relative to R-only conditioning.}
\label{tab:calibration_inventory}
\setlength{\tabcolsep}{4pt}
\renewcommand{\arraystretch}{1.20}

\begin{tabularx}{\textwidth}{@{}
>{\raggedright\arraybackslash}p{0.14\textwidth}
>{\raggedright\arraybackslash}X
>{\raggedright\arraybackslash}X
>{\raggedright\arraybackslash}X@{}}

\toprule
Development stage &
Candidates compared &
Selection criterion &
Selected result \\
\midrule

A interface design &
24 combinations of reporting systems, pressure-matching windows, and residual
representations &
Mean RMSE changes across all 13 variables and across the upper-air temperature and wind
variables &
ACARS direct, MDCRS/ARINC, and Canadian AMDAR; ${\pm}25$-hPa windows;
equal-cell mean residuals \\

S interface design &
3 residual representations &
Mean RMSE changes across all 13 variables and across the surface-targeted
variables &
Equal-cell mean residuals \\

\specialrule{0.8pt}{5pt}{5pt}

A likelihood parameters &
Staged parameter comparisons using R+A &
RMSE across the upper-air temperature and wind variables
and across all 13 variables &
$\lambda_{\mathrm A}=0.4$\newline
$\mathrm{std}_{\mathrm A}=5\times10^{-4}$\newline
$\gamma_{\mathrm A}=2\times10^{-5}$ \\

S likelihood parameters &
Staged parameter comparisons using R+S &
Surface-targeted and all-variable RMSE &
$\lambda_{\mathrm S}=0.4$\newline
$\mathrm{std}_{\mathrm S}=1.25\times10^{-4}$\newline
$\gamma_{\mathrm S}=4\times10^{-5}$ \\

\specialrule{0.8pt}{5pt}{5pt}

R+A+S selection &
9 combinations of candidate A and S likelihood-parameter settings &
Mean RMSE change across all 13 variables as the primary score, with the surface-targeted group and the
upper-air temperature and wind group as secondary checks &
The selected A and S settings above are combined, producing an $11.961\%$
mean RMSE reduction across all 13 variables in 2019 \\

\bottomrule
\end{tabularx}
\end{table}

\subsection{Interface-Design Comparisons}

The first stage determines which observations enter the A and S interfaces and
how their residuals are represented before the likelihood parameters are
calibrated. For A, the 24 candidates combine three reporting-system
selections, two pressure-matching windows, and four residual representations.
For S, the data product, quality-control criteria, and CONUS domain are already
specified, so the three candidates differ only in their residual
representations.

\begin{figure}[!htbp]
    \centering
    \includegraphics[width=0.95\textwidth]
    {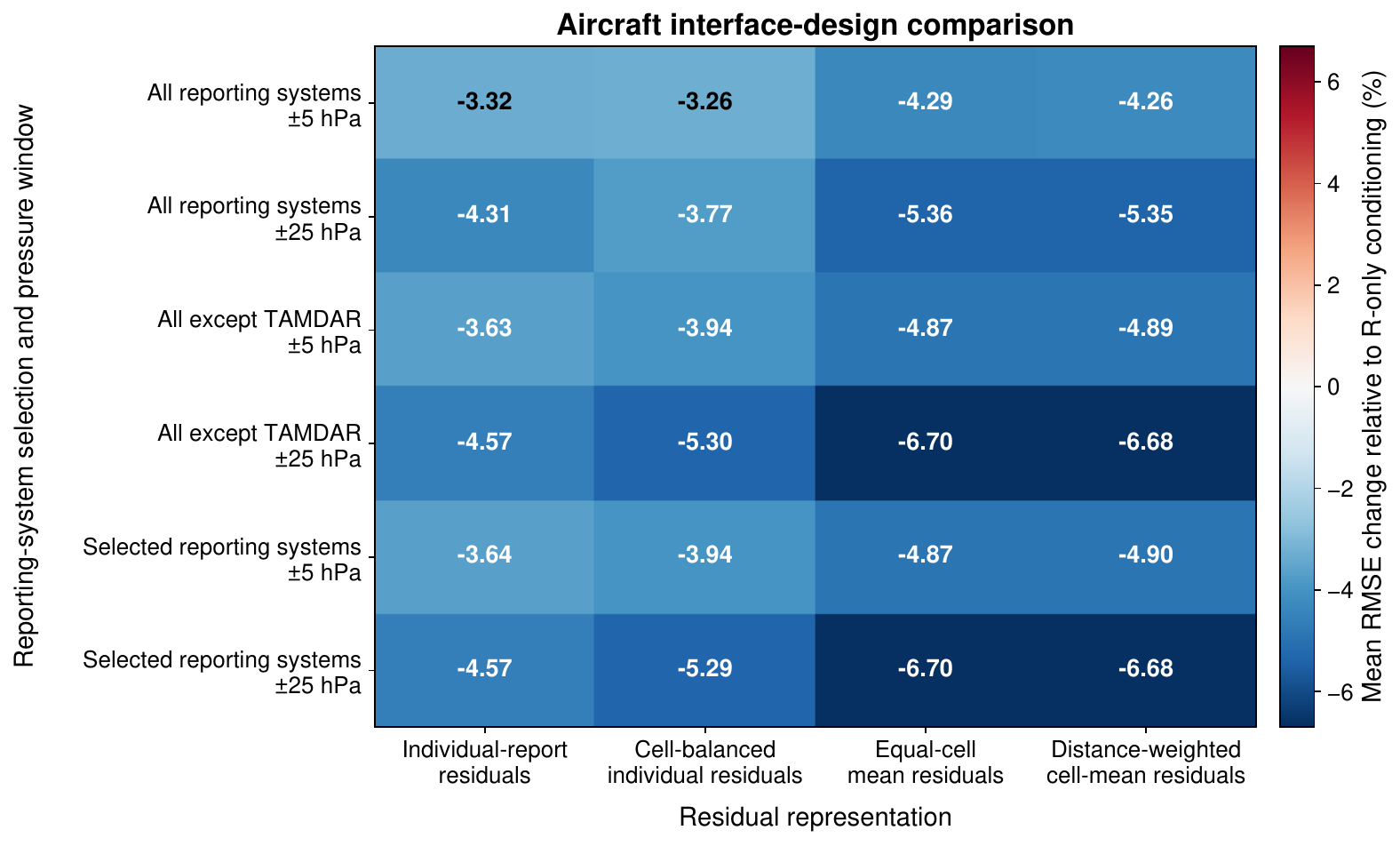}
    \caption{Aircraft interface-design comparison using the 24 prespecified
    2019 development cases. Rows compare all candidate reporting systems, all
    systems except TAMDAR, and the selected ACARS direct, MDCRS/ARINC, and
    Canadian AMDAR systems, each with $\pm5$- or $\pm25$-hPa pressure-level
    matching. Columns compare four residual representations. Values are mean
    all-variable RMSE changes relative to R-only conditioning under the same
    pre-calibration likelihood parameters. Negative values indicate lower
    RMSE. The selected design combines the selected reporting systems,
    $\pm25$-hPa matching, and equal-cell mean residuals.}
    \label{fig:abo_operator_screen}
\end{figure}

All A candidates in Figure~\ref{fig:abo_operator_screen} use the common
pre-calibration setting
$(\lambda,\mathrm{std},\gamma)=(0.1,5\times10^{-4},5\times10^{-6})$.
Holding these values fixed isolates the effects of reporting-system selection,
pressure matching, and residual representation; these are not the calibrated
parameters reported in Table~\ref{tab:final_protocols}. The selected
combination of ACARS direct, MDCRS/ARINC, and Canadian AMDAR reports and the
broader combination excluding TAMDAR produce nearly identical mean RMSE
changes over the 24 development cases. The additional non-TAMDAR systems
provide no eligible reports or only sparse coverage under the selected spatial
and pressure criteria, whereas including TAMDAR increases reconstruction
errors. We therefore select the smaller reporting-system combination together
with the ${\pm}25$-hPa windows and equal-cell mean residuals. Under the common
pre-calibration parameters, this A design reduces the mean development RMSE by
$6.697\%$ across all 13 variables and by $9.902\%$ across the
upper-air temperature and wind variables relative to R-only conditioning.

\begin{figure}[!htbp]
    \centering
    \includegraphics[width=0.86\textwidth]
    {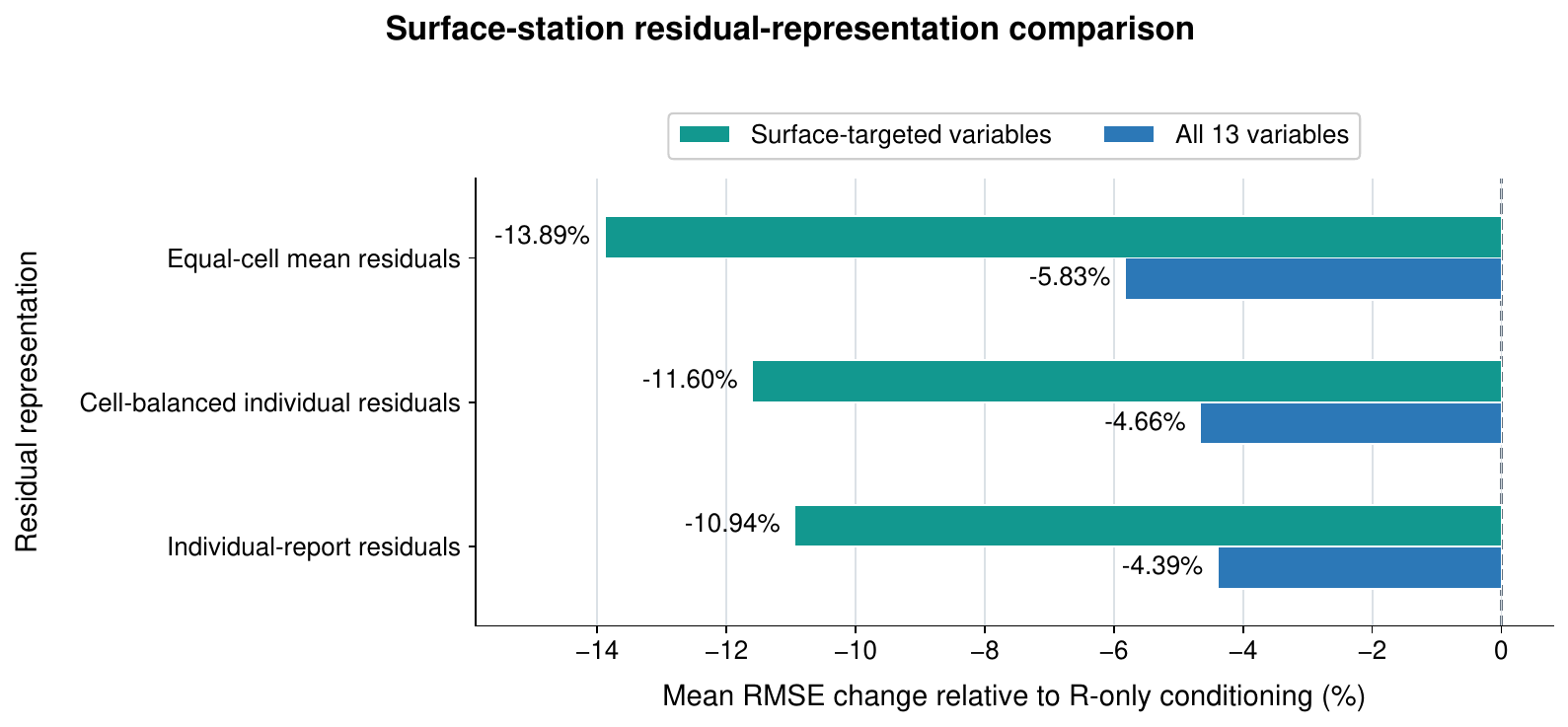}
    \caption{Surface-station residual-representation comparison using the same
    24 prespecified 2019 development cases and common pre-calibration
    likelihood parameters. The paired bars show the mean RMSE changes across
    all 13 variables and across the surface-targeted variables. Negative values
    indicate lower RMSE relative to R-only conditioning. Equal-cell mean
    residuals produce the largest reduction for both variable groups and are
    retained in the final S interface.}
    \label{fig:surface_operator_screen}
\end{figure}

Figure~\ref{fig:surface_operator_screen} compares the three S residual
representations. Equal-cell mean residuals reduce the mean development RMSE by
$5.828\%$ across all 13 variables and by $13.887\%$ across the
surface-targeted variables relative to R-only conditioning, outperforming both
individual-report residuals and cell-balanced individual residuals under the
same likelihood parameters. The cell-balanced representation retains a
separate residual for each report, whereas the equal-cell representation forms
one mean target and one residual per occupied model cell and variable. The A
and S comparisons therefore support the same design choice: report
multiplicity is controlled through residual counting before likelihood
parameters are calibrated.

\FloatBarrier

\subsection{Likelihood-Parameter Calibration and R+A+S Selection}

After selecting the interface designs, the A and S likelihood parameters are
calibrated separately using R+A and R+S, respectively. The final comparison
then evaluates nine R+A+S combinations formed from candidate A and S parameter
settings. The candidate producing the largest mean all-variable RMSE reduction
over the 24 development cases reaches $11.998\%$. Using the 10{,}000-replicate
paired-month bootstrap defined in
Section~\ref{sec:independent_year_development}, five additional candidates are
indistinguishable from this candidate because the 95\% intervals for their
paired RMSE differences include zero. Among these six candidates, the stated
selection rule chooses the setting nearest the center of the tested parameter
range, with lower combined likelihood weight used only as a final tie-breaker.
The selected R+A+S configuration combines the A and S parameters in
Table~\ref{tab:final_protocols} and produces an $11.961\%$ mean all-variable
RMSE reduction in 2019, only $0.037$ percentage points smaller than the largest
observed reduction.

\FloatBarrier

\section{Supplemental Evaluation Definitions and Diagnostics}
\label{sec:supplemental_annual}

\subsection{Metric Definitions and Aggregation}
\label{sec:metric_definitions}

The grouped and per-variable annual results use the same paired analysis-time
aggregation. For configuration $P$, analysis time $\tau$, variable $k$, grid
point $j$ in evaluation domain $G$, and ensemble member $n$, let
$x_{\tau,k,j}^{\mathrm{phys},P,(n)}$ denote the physical-unit reconstruction
and let $y_{\tau,k,j}^{\mathrm{phys}}$ denote the corresponding ERA5 reference.
The ensemble mean and its spatial RMSE are
\begin{equation}
    \overline{x}_{\tau,k,j}^{\mathrm{phys},P}
    =\frac{1}{\Nens}\sum_{n=1}^{\Nens}
    x_{\tau,k,j}^{\mathrm{phys},P,(n)},
    \qquad
    \operatorname{RMSE}_{\tau,k,G}^{P}
    =\left[
    \frac{1}{|G|}\sum_{j\in G}
    \left(\overline{x}_{\tau,k,j}^{\mathrm{phys},P}
    -y_{\tau,k,j}^{\mathrm{phys}}\right)^2
    \right]^{1/2}.
    \label{eq:ensemble_mean_rmse}
\end{equation}
The RMSE change at each analysis time is defined in
Equation~\eqref{eq:percent_delta}. For a single variable, setting
$\mathcal K'=\{k\}$ in Equation~\eqref{eq:headline_mean_of_ratios} gives the
per-variable summary $\Delta_{k,G}^{P}$. The annual-mean RMSE values in physical
units are reported separately in Table~\ref{tab:absolute_rmse_full723}.

The remaining definitions support the probabilistic diagnostics. For the same
indices, the physical-unit RMS ensemble spread uses the sample variance across
the $\Nens=16$ members:
\begin{equation}
    \operatorname{Spread}_{\tau,k,G}^{P}
    =
    \left[
    \frac{1}{|G|}
    \sum_{j\in G}\operatorname{Var}_{n=1:\Nens}
    (x_{\tau,k,j}^{\mathrm{phys},P,(n)})
    \right]^{1/2}.
    \label{eq:ensemble_spread}
\end{equation}
For a physical-unit scalar ensemble
$x^{\mathrm{phys},(1:\Nens)}$ and scalar reference $y^{\mathrm{phys}}$, the
empirical finite-ensemble CRPS is
\begin{equation}
    \operatorname{CRPS}
    =
    \frac{1}{\Nens}\sum_{n=1}^{\Nens}
    |x^{\mathrm{phys},(n)}-y^{\mathrm{phys}}|
    -
    \frac{1}{2\Nens^2}
    \sum_{n=1}^{\Nens}\sum_{n'=1}^{\Nens}
    |x^{\mathrm{phys},(n)}-x^{\mathrm{phys},(n')}|.
    \label{eq:crps}
\end{equation}
As a sensitivity check, the fair finite-ensemble estimator replaces
$\Nens^2$ in the second denominator by $\Nens(\Nens-1)$, correcting the
finite-ensemble contribution of the pairwise term under exchangeable sampling
\citep{ferro2014fair}:
\begin{equation}
    \operatorname{CRPS}_{\mathrm{fair}}
    =
    \frac{1}{\Nens}\sum_{n=1}^{\Nens}
    |x^{\mathrm{phys},(n)}-y^{\mathrm{phys}}|
    -
    \frac{1}{2\Nens(\Nens-1)}
    \sum_{n=1}^{\Nens}\sum_{n'=1}^{\Nens}
    |x^{\mathrm{phys},(n)}-x^{\mathrm{phys},(n')}|.
    \label{eq:fair_crps}
\end{equation}
At each analysis time, the pointwise scores are averaged over the evaluation
domain before configurations are compared:
\begin{equation}
\begin{aligned}
    \operatorname{CRPS}_{\tau,k,G}^{P}
    &=\frac{1}{|G|}\sum_{j\in G}
    \operatorname{CRPS}\!\left(
    x_{\tau,k,j}^{\mathrm{phys},P,(1:\Nens)},
    y_{\tau,k,j}^{\mathrm{phys}}\right),\\
    \operatorname{CRPS}_{\mathrm{fair},\tau,k,G}^{P}
    &=\frac{1}{|G|}\sum_{j\in G}
    \operatorname{CRPS}_{\mathrm{fair}}\!\left(
    x_{\tau,k,j}^{\mathrm{phys},P,(1:\Nens)},
    y_{\tau,k,j}^{\mathrm{phys}}\right).
\end{aligned}
\label{eq:domain_mean_crps}
\end{equation}
The finite-ensemble-corrected spread--skill ratio over the matched analysis-time
set $\mathcal T$ is
\begin{equation}
    \operatorname{SSR}_{k,G}^{P}
    =
    \sqrt{\frac{\Nens+1}{\Nens}}
    \left[
    \frac{\sum_{\tau\in\mathcal T}
    (\operatorname{Spread}_{\tau,k,G}^{P})^{2}}
         {\sum_{\tau\in\mathcal T}
    (\operatorname{RMSE}_{\tau,k,G}^{P})^{2}}
    \right]^{1/2}.
    \label{eq:spread_skill_ratio}
\end{equation}
The correction accounts for the additional sampling variance of the truth
relative to an $\Nens$-member ensemble mean; exchangeability gives a reference
value of one, while values below one indicate insufficient spread relative to
ensemble-mean RMSE \citep{fortin2014spread}.

The grouped CRPS comparison uses the same paired analysis-time aggregation as
the primary grouped RMSE comparison:
\begin{equation}
    \overline{\Delta}^{P,\mathrm{CRPS}}_{G,\mathcal K'}
    =\frac{100}{|\mathcal T|\,|\mathcal K'|}
    \sum_{\tau\in\mathcal T}\sum_{k\in\mathcal K'}
    \frac{\operatorname{CRPS}_{\tau,k,G}^{P}
    -\operatorname{CRPS}_{\tau,k,G}^{\mathrm{R}}}
    {\operatorname{CRPS}_{\tau,k,G}^{\mathrm{R}}}.
    \label{eq:crps_mean_of_ratios}
\end{equation}
Thus, neither grouped RMSE nor grouped CRPS averages incompatible physical
units. Coverage is already dimensionless and is averaged over the same matched
combinations of analysis times and variables.

\begin{table}[!htbp]
    \centering
    \caption{Physical-unit annual-mean RMSE against ERA5 over the CONUS domain
    across 723 matched analysis times. Humidity RMSE is shown in
    $10^{-3}\,\mathrm{kg\,kg^{-1}}$.}
    \label{tab:absolute_rmse_full723}
    \small
    \renewcommand{\arraystretch}{1.10}
    \begin{tabular*}{\textwidth}{@{\extracolsep{\fill}}llrr@{}}
        \toprule
        Variable & Unit & R-only RMSE & R+A+S RMSE \\
        \midrule
        $t2m$ & $\mathrm{K}$ & 1.725 & 1.380 \\
        $u10$ & $\mathrm{m\,s^{-1}}$ & 1.605 & 1.414 \\
        $v10$ & $\mathrm{m\,s^{-1}}$ & 1.697 & 1.507 \\
        $z500$ & $\mathrm{m^2\,s^{-2}}$ & 112.4 & 101.8 \\
        $z850$ & $\mathrm{m^2\,s^{-2}}$ & 84.5 & 76.1 \\
        $u500$ & $\mathrm{m\,s^{-1}}$ & 2.912 & 2.618 \\
        $u850$ & $\mathrm{m\,s^{-1}}$ & 2.404 & 2.157 \\
        $v500$ & $\mathrm{m\,s^{-1}}$ & 3.042 & 2.695 \\
        $v850$ & $\mathrm{m\,s^{-1}}$ & 2.539 & 2.287 \\
        $t500$ & $\mathrm{K}$ & 0.901 & 0.861 \\
        $t850$ & $\mathrm{K}$ & 1.396 & 1.242 \\
        $q500$ & $10^{-3}\,\mathrm{kg\,kg^{-1}}$ & 0.513 & 0.506 \\
        $q850$ & $10^{-3}\,\mathrm{kg\,kg^{-1}}$ & 1.286 & 1.263 \\
        \bottomrule
    \end{tabular*}
\end{table}

\subsection{Additional Temporal-Uncertainty Results}

The temporal-uncertainty analysis in the main text describes the 14-day
moving-block bootstrap and reports the grouped intervals. For the per-variable
intervals, this bootstrap samples circular blocks of 28 consecutive
analysis-time positions from the complete 2020 00/12 UTC calendar. The nine unavailable
analysis times remain calendar gaps rather than being removed before
resampling.

For each variable, this resampling is applied to
$\Delta_{\tau,k,G}^{P}$ as defined in Equation~\eqref{eq:percent_delta}. The
reported mean is the per-variable summary $\Delta_{k,G}^{P}$.

\begin{figure}[!htbp]
    \centering
    \includegraphics[width=0.92\textwidth]
    {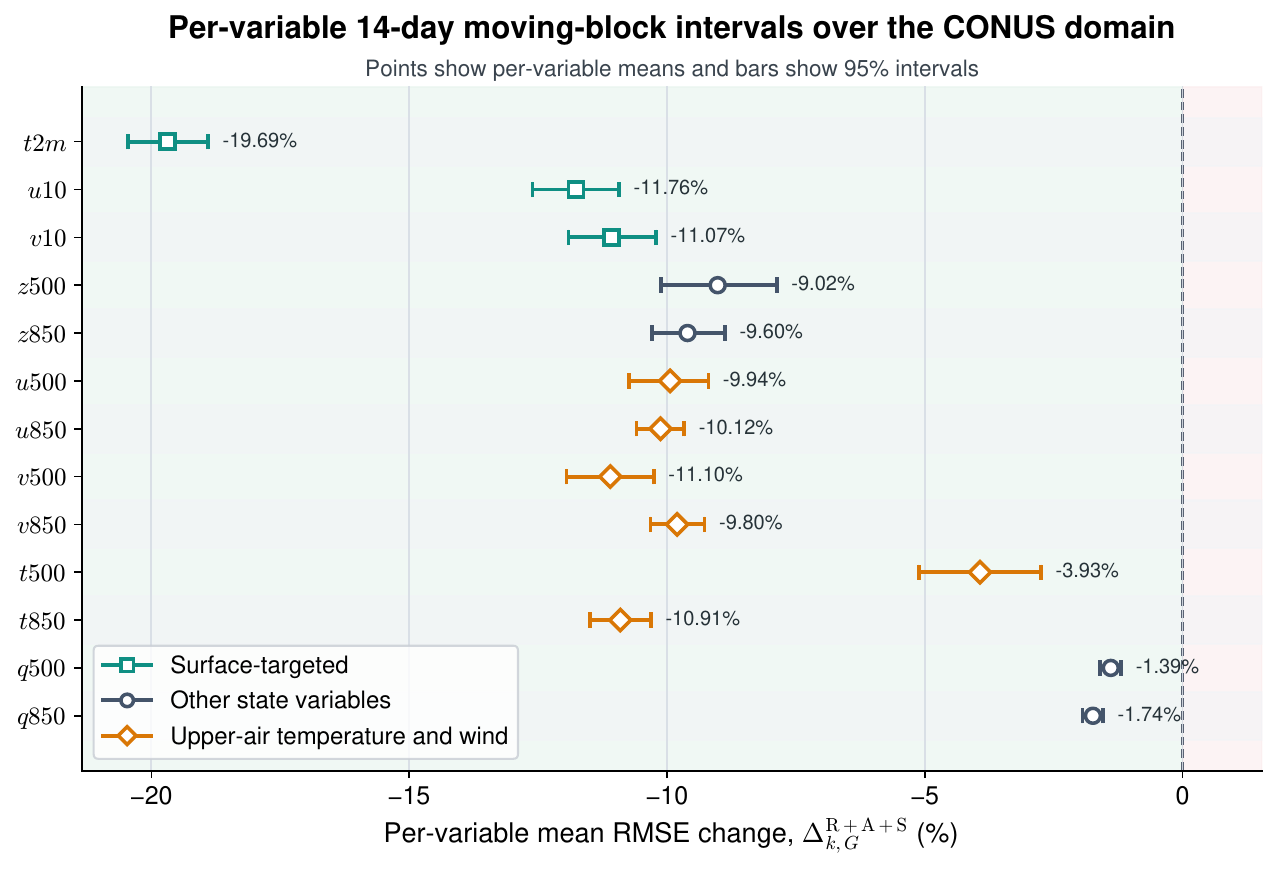}
    \caption{Per-variable mean RMSE changes
    $\Delta_{k,G}^{\mathrm{R+A+S}}$ over the CONUS domain, with 95\% intervals
    from the 14-day moving-block bootstrap. Negative values indicate lower
    RMSE relative to R-only conditioning.}
    \label{fig:frozen2019_per_variable_bootstrap}
\end{figure}

Table~\ref{tab:per_variable_bootstrap_final} gives exact per-variable
intervals for R+A+S relative to R-only conditioning, and
Figure~\ref{fig:frozen2019_per_variable_bootstrap} shows their pattern across
the state. Marker shape and color distinguish variables directly constrained
by S, variables directly constrained by A, and variables not directly
constrained by either source. All 13 intervals lie below zero.

\begin{table}[!htbp]
\centering
\small
\caption{Per-variable mean RMSE changes
$\Delta_{k,G}^{\mathrm{R+A+S}}$ over the CONUS domain. The 95\% intervals are
computed using the 14-day moving-block bootstrap. Negative values indicate
lower RMSE relative to R-only conditioning.}
\label{tab:per_variable_bootstrap_final}
\renewcommand{\arraystretch}{1.08}
\begin{tabular*}{\textwidth}{@{\extracolsep{\fill}}lrrlrr@{}}
\toprule
Variable & $\Delta_{k,G}^{\mathrm{R+A+S}}$ (\%) & 95\% interval (\%) &
Variable & $\Delta_{k,G}^{\mathrm{R+A+S}}$ (\%) & 95\% interval (\%) \\
\midrule
$t2m$  & $-19.686$ & $[-20.452,-18.901]$ &
$u10$  & $-11.760$ & $[-12.606,-10.929]$ \\
$v10$  & $-11.072$ & $[-11.913,-10.211]$ &
$z500$ & $-9.017$  & $[-10.112,-7.869]$ \\
$z850$ & $-9.602$  & $[-10.291,-8.876]$ &
$u500$ & $-9.940$  & $[-10.734,-9.199]$ \\
$u850$ & $-10.124$ & $[-10.595,-9.667]$ &
$v500$ & $-11.099$ & $[-11.950,-10.256]$ \\
$v850$ & $-9.805$  & $[-10.319,-9.269]$ &
$t500$ & $-3.930$  & $[-5.115,-2.745]$ \\
$t850$ & $-10.905$ & $[-11.488,-10.311]$ &
$q500$ & $-1.392$  & $[-1.602,-1.191]$ \\
$q850$ & $-1.739$  & $[-1.937,-1.541]$ &
& & \\
\bottomrule
\end{tabular*}
\end{table}

\FloatBarrier

\subsection{Spatial Spectral Shape}
\label{sec:supplemental_spatial_fields}

\begin{figure}[!htbp]
    \centering
    \includegraphics[width=0.99\textwidth]
    {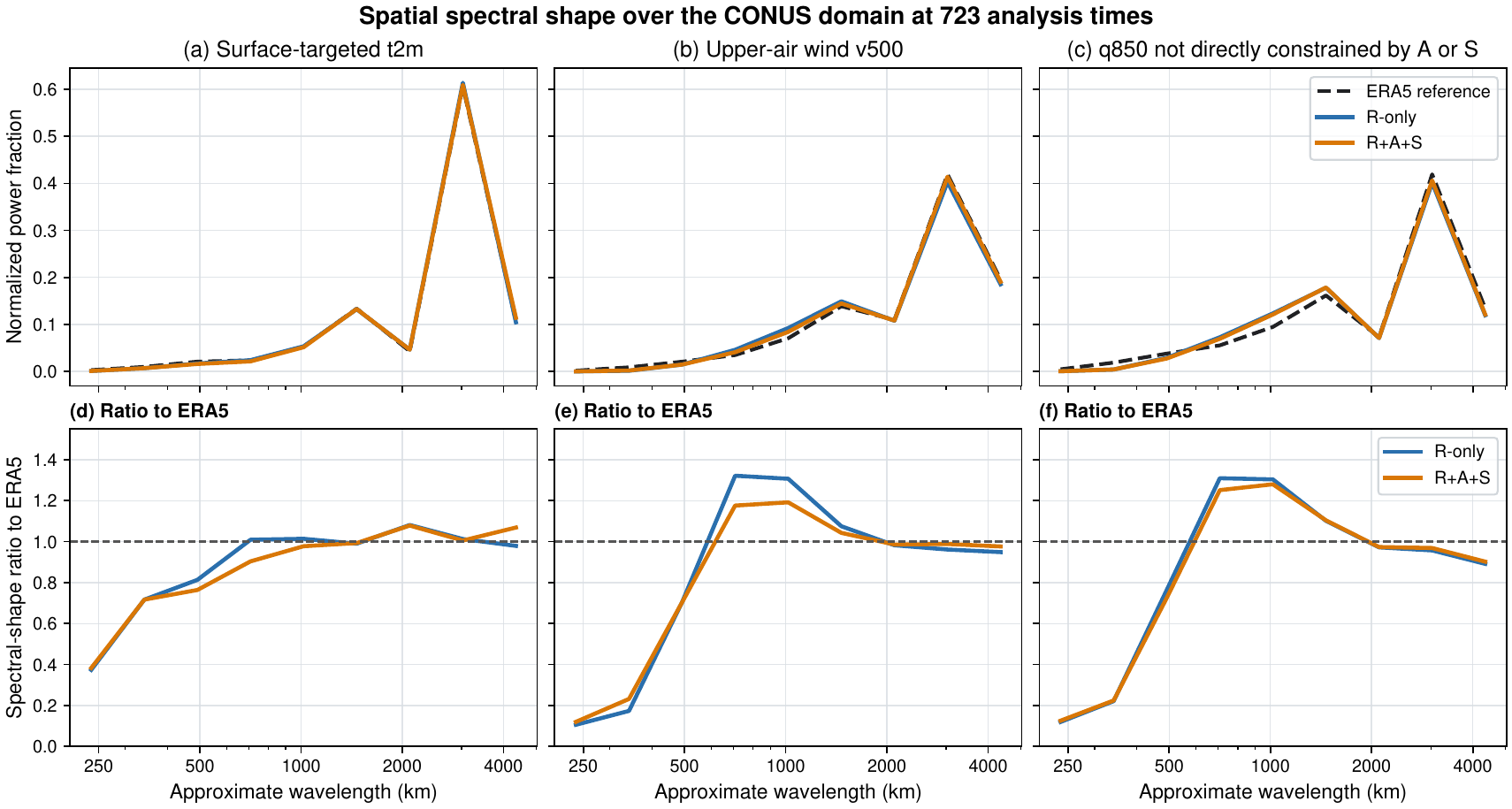}
    \caption{Spatial spectra over the CONUS domain at 723 analysis times.
    Panels (a)--(c) show mean normalized power fractions for a
    surface-targeted variable, an upper-air wind variable, and a variable
    not directly constrained by A or S. Panels (d)--(f) show the corresponding
    ratios to ERA5.}
    \label{fig:spatial_spectral_shape}
\end{figure}

Complementing the selected field examples in the main text, this diagnostic
tests whether the lower RMSE of R+A+S is accompanied by uniform spectral
smoothing over the 723 evaluation times. Let $\Omega$ denote the CONUS-domain
grid. For each analysis time $\tau$ and variable $k$, let $z^P_{\tau,k}$ denote
the 16-member ensemble-mean field for
$P\in\{\mathrm{R},\mathrm{R{+}A{+}S}\}$ and the deterministic reference field
for $P=\mathrm{ERA5}$. After removing the domain mean and applying a separable
Hann window $w$, we compute the normalized power in radial-wavenumber bin
$B_b$ as
\begin{equation}
 p^P_{\tau,k,b}
 =
 \frac{\displaystyle\sum_{\boldsymbol{\kappa}\in B_b}
 \left|\mathcal{F}\!\left\{w\left(z^P_{\tau,k}
 -\langle z^P_{\tau,k}\rangle_{\Omega}\right)\right\}
 (\boldsymbol{\kappa})\right|^2}
 {\displaystyle\sum_{\boldsymbol{\kappa}\ne\boldsymbol{0}}
 \left|\mathcal{F}\!\left\{w\left(z^P_{\tau,k}
 -\langle z^P_{\tau,k}\rangle_{\Omega}\right)\right\}
 (\boldsymbol{\kappa})\right|^2}.
 \label{eq:normalized_spatial_spectrum}
\end{equation}
The nine logarithmically spaced bins span the nonzero discrete wavenumbers;
their displayed wavelengths are the inverse geometric bin centers. We first
normalize each analysis time as in
Eq.~\eqref{eq:normalized_spatial_spectrum}, then average over the matched set
$\mathcal T$:
\begin{equation}
    \bar p^P_{k,b}
    =\frac{1}{|\mathcal T|}\sum_{\tau\in\mathcal T}p^P_{\tau,k,b},
    \qquad |\mathcal T|=723.
    \label{eq:mean_normalized_spatial_spectrum}
\end{equation}
The lower panels report ratios of these temporal means,
$\bar p^P_{k,b}/\bar p^{\mathrm{ERA5}}_{k,b}$, rather than temporal means of
per-analysis-time ratios. Figure~\ref{fig:spatial_spectral_shape} reports both
$\bar p^P_{k,b}$ and these ratios. R+A+S leaves the broad spectral shape close
to R-only for $t2m$ and $q850$, while for $v500$ it moves several intermediate and
long scales toward ERA5. Across these examples, R+A+S does not uniformly
suppress relative power across spatial scales. Because each field is
normalized by its total nonzero power, this diagnostic evaluates relative
spectral shape rather than absolute variance or phase accuracy. The limited
size of the CONUS-domain grid also restricts interpretation near the grid
scale.

\FloatBarrier

\section{Supplemental Probabilistic Diagnostics}
\label{sec:supplemental_validation}

\subsection{Finite-Ensemble Coverage Interpretation}
\label{sec:finite_ensemble_coverage}

The empirical coverage diagnostic uses linearly interpolated 5th and 95th
percentiles estimated from the 16 ensemble members. Because these percentiles
are estimated from a finite ensemble, 0.90 is a large-ensemble reference rather
than the exact expected coverage for this experiment. Under exchangeability,
even the wider interval bounded by the sample minimum and maximum contains an
independently exchangeable verifying value with probability
\begin{equation*}
    \frac{\Nens-1}{\Nens+1}
    =\frac{15}{17}
    \approx0.882.
\end{equation*}
The interpolated 5th--95th percentile interval lies within this envelope.
We therefore interpret its empirical coverage comparatively across
conditioning configurations and together with the spread--skill ratios and
rank frequencies, rather than as a stand-alone calibration test.

\subsection{Annual Probabilistic Diagnostics}

\begin{figure}[!htbp]
    \centering
    \includegraphics[width=0.98\textwidth]
    {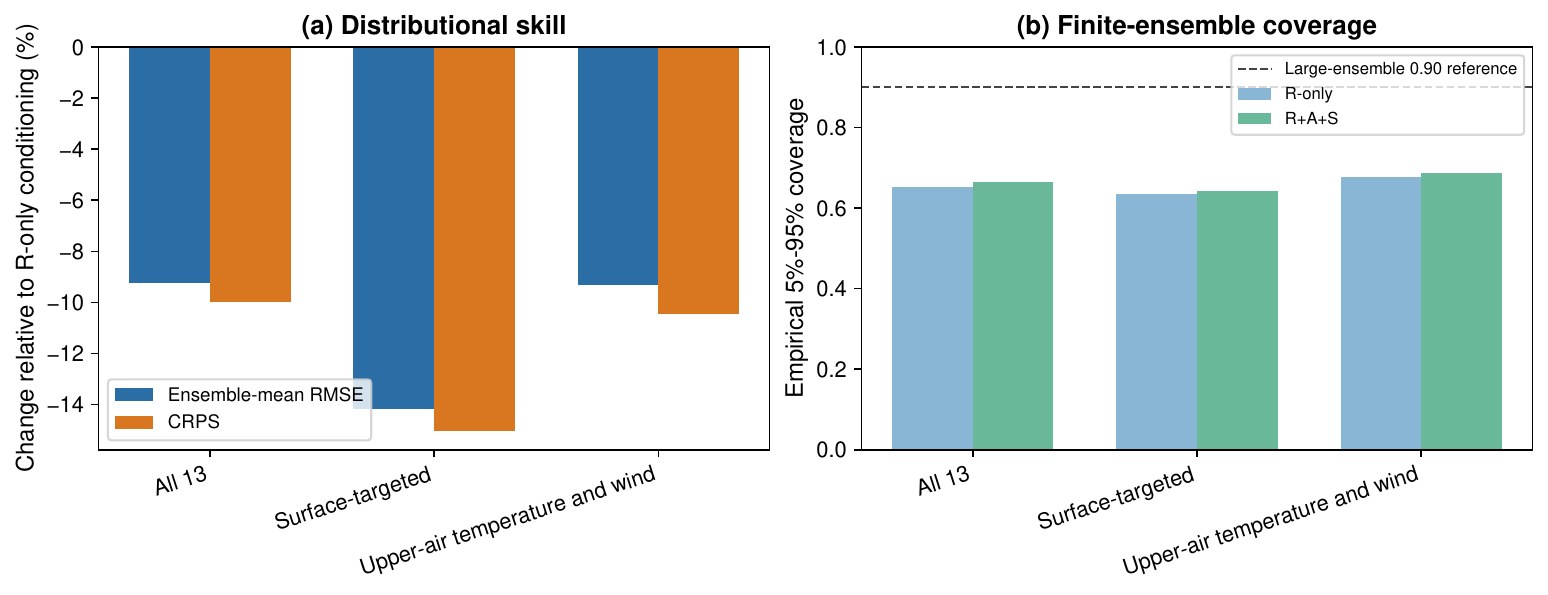}
    \caption{Full-year probabilistic diagnostics over the CONUS domain for
    R-only and R+A+S conditioning: (a) changes in CRPS and ensemble-mean RMSE
    relative to R-only conditioning and (b) empirical coverage of the
    5th--95th percentile ensemble interval.}
    \label{fig:frozen2019_probabilistic}
\end{figure}

Figure~\ref{fig:frozen2019_probabilistic} compares the full-year changes in
CRPS and ensemble-mean RMSE and reports the corresponding finite-ensemble
coverage. R+A+S lowers both CRPS and ensemble-mean RMSE relative to R-only
conditioning, while the coverage values remain below the large-ensemble 0.90
reference. The fair-CRPS sensitivity defined in Eq.~\eqref{eq:fair_crps} uses
the same 16 members and leaves the ordering of the reported configurations
unchanged.

\FloatBarrier

\begin{figure}[!htbp]
    \centering
    \includegraphics[width=0.98\textwidth]
    {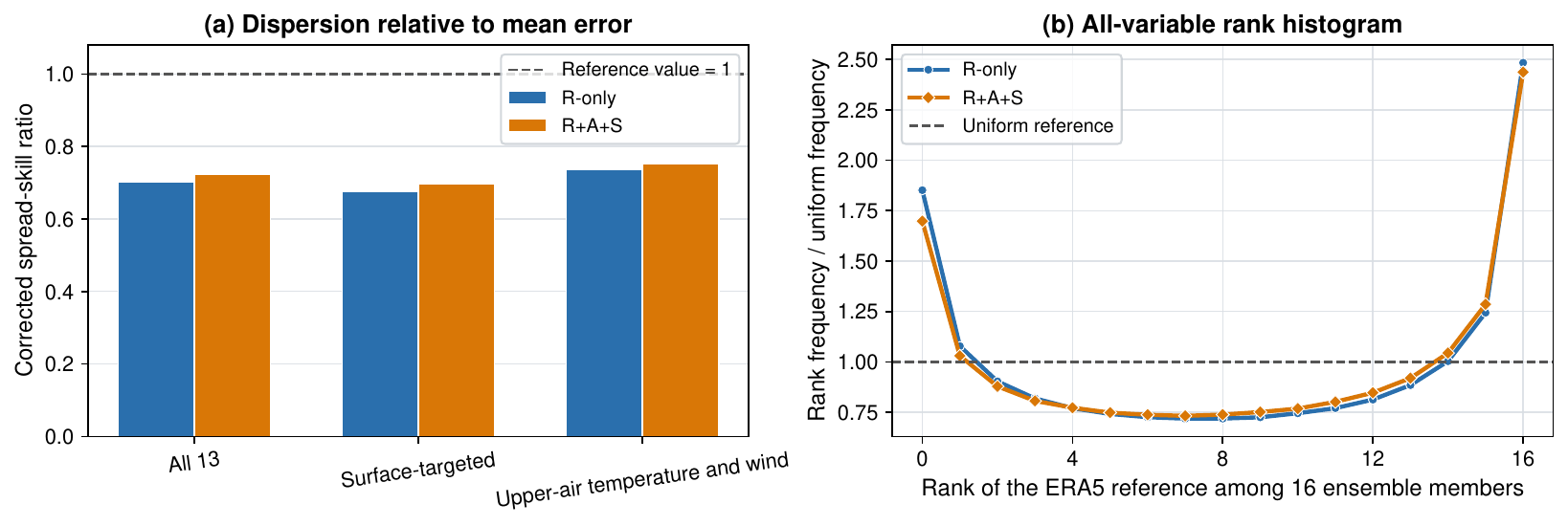}
    \caption{Ensemble-dispersion diagnostics over the CONUS domain at 723
    matched 2020 analysis times: (a) finite-ensemble-corrected spread--skill
    ratios and (b) all-variable rank frequencies normalized by the frequency
    expected under a uniform rank distribution.}
    \label{fig:frozen2019_spread_skill_rank}
\end{figure}

Figure~\ref{fig:frozen2019_spread_skill_rank} examines ensemble dispersion
using the finite-ensemble-corrected spread--skill ratio and pooled rank
frequencies. The spread--skill ratios remain below the reference value of one
for both configurations, and the rank frequencies are elevated near the outer
ranks. Together, these diagnostics indicate that the ensembles remain
underdispersed even though R+A+S improves CRPS and ensemble-mean RMSE.

Table~\ref{tab:frozen2019_spread_skill_variable} reports the corresponding
spread--skill ratios separately for each state variable.

\begin{table}[!htbp]
    \centering
    \caption{Finite-ensemble-corrected spread--skill ratios by state variable
    over the CONUS domain at 723 matched 2020 analysis times.}
    \label{tab:frozen2019_spread_skill_variable}
    \small
    \renewcommand{\arraystretch}{1.08}
    \begin{tabular*}{\textwidth}{@{\extracolsep{\fill}}lcclcc@{}}
        \toprule
        Variable & R-only & R+A+S & Variable & R-only & R+A+S \\
        \midrule
        $t2m$  & 0.721 & 0.793 & $u500$ & 0.752 & 0.768 \\
        $u10$  & 0.667 & 0.664 & $u850$ & 0.701 & 0.709 \\
        $v10$  & 0.642 & 0.631 & $v500$ & 0.704 & 0.724 \\
        $z500$ & 0.577 & 0.639 & $v850$ & 0.677 & 0.678 \\
        $z850$ & 0.613 & 0.680 & $t500$ & 0.856 & 0.858 \\
        $q500$ & 0.679 & 0.682 & $t850$ & 0.734 & 0.777 \\
        $q850$ & 0.793 & 0.795 & & & \\
        \bottomrule
    \end{tabular*}
\end{table}

\end{document}